\documentclass[11pt]{article}

\usepackage[margin=1in]{geometry}
\usepackage[utf8]{inputenc}
\usepackage[T1]{fontenc}
\usepackage{amsmath,amssymb,amsthm}
\usepackage{bm}
\usepackage{mathtools}
\usepackage{graphicx}
\usepackage{booktabs}
\usepackage{multirow}
\usepackage{algorithm}
\usepackage{algorithmic}
\usepackage[colorlinks=true,citecolor=blue,linkcolor=blue,urlcolor=blue]{hyperref}
\usepackage[round,authoryear]{natbib}   
\usepackage{enumitem}
\usepackage{xcolor}
\usepackage{authblk}                    

\theoremstyle{definition}
\newtheorem{definition}{Definition}

\newtheorem{remark}{Remark}

\title{\Large\bfseries Neural Feature Governance: Extending Atom Prevalence}

\author[1]{Idris Karel Seunda Ekwe\thanks{Corresponding author. Email: \texttt{idris.seunda@aims-cameroon.org}}}
\author[1]{Patrick Tenga Shako\thanks{Email: \texttt{patrick.tenga@aims-cameroon.org}}}
\author[2]{Ernest Parfait Fokou\'e\thanks{Corresponding author. Email: \texttt{epfeqa@rit.edu}}}

\affil[1]{African Institute for Mathematical Sciences (AIMS), Cameroon}
\affil[2]{Rochester Institute of Technology (RIT), US}

\date{}

\begin{document}

\maketitle

\begin{abstract}
\noindent Neural network compression and interpretability remain open challenges in modern deep learning,
where billion-parameter architectures deliver impressive accuracy at the cost of transparency,
computational efficiency, and reliable uncertainty quantification. This paper introduces
Neural Atom Prevalence (NAP), a principled Bayesian framework for structured node-level
model selection in feedforward neural networks. NAP introduces the neural atom (activation unit) and functions as a hybrid method operating through a four-phase
pipeline: Bayesian Lottery Ticket (BLT) identification via Iterative Magnitude Pruning (IMP), soft
variational training of the Spike and Slab Independent Gaussian (SS-IG) model, Poisson-Binomial (PB)
optimal layer-size selection, and Bayesian fine-tuning to produce a sparse, stable, interpretable, and accurate model.
 Extensive empirical validation across simulated
nonlinear regression, two UCI benchmark datasets (Concrete, YearPredictionMSD), and the MNIST
image classification task demonstrates that NAP achieves state-of-the-art structural sparsity,
reducing active nodes to as few as $8\%$ of the original dense architecture on MNIST, while
well-calibrated probabilistically: the aleatoric-epistemic uncertainty decomposition reveals
that model ignorance accounts for only 3 to $4\%$ of total predictive variance across all
experiments, and regression reliability diagrams confirm a near-nominal predictive interval
coverage (93.4\% observed against a 95\% target). These results establish NAP as a reliable,
theoretically grounded, and computationally tractable solution to the simultaneous pursuit of
sparsity, accuracy, interpretability, and uncertainty quantification in Bayesian neural networks.
\end{abstract}


\section{Introduction}
\label{sec:introduction}

The neural network is now gaining important interest in the machine learning environment due to their high flexibility and its capacity to approximate any function with high precision (\cite{hornik1989multilayer}). However, if having a model that achieves higher performance is great, understanding how it achieves this performance is crucial. The Bayesian approach is known as an appropriate framework for construction of interpretable Deep learning models that allow quantification of uncertainty (\cite{blundell2015weight}). But neural network models are characterized by a low level of interpretability (usually called a black box) and their complexity, related to their architecture, sometimes composed of networks of billions of weights and neurons which make it difficult to implement. Additionally,  although the Bayesian framework offers uncertainty quantification, it unfortunately has the drawback of making calculations heavier, thereby increasing computational performance demands for building models that are efficient, flexible, interpretable, and stable (\cite{blundell2015weight}). To overcome these issues, the model compression has been found as a good strategy, trying to reduce model sizes and complexity while keeping acceptable performances, interpretability and stability. Model compression can be achieve by improving the sparsity of models, feature selection or more generally model selection. All these include the same concepts of, going from a given set of variables, atoms or neurons, eliminate those with nonsignificant or not positive contribution to the success of final machine learning task (\cite{han2015learning, jantre2022layer}).

Today, research on interpretable and sparse neural network architectures has achieved great progress with the development of an empirical framework such as the Lottery ticket hypothesis (\cite{frankle2019lottery}) able to prune up to 80\% of the network with at least the same performance as the dense network. This frame has been converted into Bayesian to develop the Bayesian winning ticket (\cite{kuhn2026bayesian}) for uncertainty quantification. However, it performs at the weights level and is characterized by an unstructured pruning strategy (\cite{han2015learning, frankle2019lottery}) or magnitude pruning, which rely more on the value of the weight than the probability to be included in the final model. 

More robust and theoretical frame has been developed, performing model compression at the neuron level and playing on the choice of the prior distribution of model parameters to force some neurons to be silent or completely pruned during training, allowing the model size to decrease progressively during the training (\cite{jantre2022layer}). But the compression level of the network is sometimes very small compared to the Bayesian lottery framework (\cite{jantre2022layer, kuhn2026bayesian}). Also, premature pruning can in some cases result in a loss of accuracy due to nodes removed in earlier stages of the training that could contribute considerably by the end of the training. These limitations highlight the need of a unified framework, which combines both the ability to compress efficiently by using the Bayesian Lottery Ticket approach, as well as the inclusion of neurons probabilistically, which results in efficient sparse networks. 

The main objective of this thesis is to develop a neural atom prevalence method for neural network model construction as an answer to the question: Can Fokou\'e's atom prevalence principle~(\cite{fokoue2008estimation}), based on identifying the most likely model size and selecting the top-ranked atoms, be successfully extended to neural networks under a variational Bayesian framework?. More precisely, it aims at defining a framework for neural network model construction through neuron selection that achieves a benchmark network compression level, preserving or improving model performance and interpretability.  To achieve this goal, we define the neural atom; Construct the theoretical framework of the neural atom prevalence method; Leverage the performance of Bayesian Lottery ticket and the Spike and Slab Independent Gaussian (SS-IG) prior into a single unified frame for Bayesian neural network model construction; and validate the method empirically.

The following part of the paper is organized into four sections. Section \ref{sec:related-work} reviews the related work. Section \ref{sec:methodology} presents the proposed neural atom prevalence method; Section \ref{sec:experiments} presents the experiments and empirical validation of the method and  the final Section \ref{sec:conclusion} is the conclusion.

\section{Related Work}
\label{sec:related-work}

A model that cannot explain which inputs drove its predictions provides limited value to scientists, clinicians, or regulators who must justify decisions. The goal of feature selection or model selection is therefore not only predictive accuracy but also identifying a sparse or compressed, stable, and interpretable subset of inputs or model components that are really necessary for prediction. The following paragraphs review the flow of feature selection in Bayesian frame work, from the linear models to neural network models, and then anounce the paper proposed method.

In the area of feature selection, one of the key contributions in the Bayesian framework, and the direct theoretical foundation of this paper, is the Atom Prevalence Method of Fokou\'{e}~\cite{fokoue2008estimation}.
It came directly upon the limitation of the Median Probability Method (MP)~\cite{barbieri2004optimal}, which aims to identify the optimal model among a set of possible models, built on a set of atoms assimilated to features. The MP method is based on a cut-off posterior inclusion probability $p=\frac{1}{2}$ in feature selection, which was successful in many cases but inefficient when no posterior inclusion was equal to or above $\frac{1}{2}$. Fokou\'{e} solved this problem by introducing an adaptive cut-off to guarantee the existence of the optimal model. More specifically, he synthesized the exact sparsity of Mitchell and Beauchamp with the computational tractability of SSVS by using birth-and-death MCMC, a continuous-time Markov process that adds (birth) or removes (death) variables from the active set without discrete accept/reject steps. Under the full hierarchical model:
\begin{align*}
	\beta_i \mid \lambda_i &\sim \mathcal{N}(\beta_i;\, 0,\, \lambda_i^{-1}), 
	\quad \lambda_i \mid a, b \sim \text{Gamma}(\lambda_i;\, a, b), \quad k \sim \text{Poisson}(\omega)\cdot\mathbf{1}[1 \leq k \leq m],\\
	v_i \mid \pi &\sim \text{Bernoulli}(\pi), \quad \pi \sim \text{Beta}(a_\pi, b_\pi),\quad \sigma^{-2} \mid c, d \sim \text{Gamma}(\sigma^{-2};\, c, d)
\end{align*}
the birth rate for an inactive variable $j$ given current active set $A$ is proportional to the Bayes factor $p(\mathbf{y} \mid A \cup \{j\})/p(\mathbf{y} \mid A)$, and the death rate proportionally to its reciprocal.
	
From the resulting samples, three key quantities are computed: (1) the inclusion probability $\hat{p}_j = \Pr(v_j = 1 \mid \mathbf{y})$ for each predictor; (2) the optimal model size $k^{\mathrm{opt}} = \arg\max_k \Pr(|A|=k \mid \mathbf{y})$, used find the adaptive cut off; and (3) the selected model, comprising the $k^{\mathrm{opt}}$ predictors with highest $\hat{p}_j$. Atom Prevalence thereby delivers exact Bayesian sparsity, full uncertainty quantification, and scalability to $p$, addressing the limitations of the MP model and three preceding approaches.
	
The above method is discussed for linear models (The atoms are linear predictors $x_j$), putting aside the large family of non-linear models, among which are neural networks. In a neural network, the natural atoms are neuron activations $h_j(\mathbf{x})$, the non-linear, learned representations. Extending Atom Prevalence from predictors to neurons is the central contribution of this thesis.
	
Neural networks violate the premise of linear variable selection: there is no fixed, interpretable set of features to select, and the parameter count vastly exceeds the sample size. The canonical response to over-parameterisation has been pruning, the removal of redundant weights or neurons after training. LeCun, Denker, and Solla (\cite{lecun1989optimal}) first formalised this via a second-order Taylor expansion of the training loss, removing weights whose removal caused the smallest increase in loss (Optimal Brain Damage). Han et al. (\cite{han2015learning}) demonstrated the same principle at scale: magnitude pruning alone removes over 90\% of weights from large convolutional networks without accuracy loss. These results empirically establish that trained networks are massively over-parameterised, but they provide no probabilistic characterization of which weights or neurons are structurally necessary.

Frankle and Carlin~(\cite{frankle2019lottery}) transformed the pruning narrative with the Lottery Ticket Hypothesis (LTH): a randomly-initialised dense network contains a sparse subnetwork  (a winning ticket) that, when trained in isolation from the original initialisation $\theta_0$, matches the full network's accuracy in at most the same number of iterations. The identification procedure is iterative magnitude pruning with weight resetting:
	\begin{enumerate}
		\item Randomly initialise $f(\mathbf{x};\,\theta_0)$ and train to convergence, obtaining $\theta_T$.
		\item Prune the $p\%$ smallest-magnitude weights, creating mask $m$.
		\item Reset surviving weights to $\theta_0$ (not $\theta_T$).
		\item Repeat until target sparsity is reached.
	\end{enumerate}
	The decisive control experiment is random reinitialization: the winning ticket's mask $m$ applied to a fresh random initialisation $\theta_0'$ performs no better than random pruning, proving that the specific initial values $m \odot \theta_0$ (not merely the sparse architecture) constitute the ticket.
	
	The LTH has two implications for this thesis. First, it establishes that sparse subnetworks with specific structural properties exist a priori, a finding that is structurally analogous to the spike-and-slab assertion that a sparse subset of predictors is active. Second, it shows that which neurons survive matters, not just how many. Moving from a frequentist to probabilistic frame, Kuhn et al.\ (\cite{kuhn2026bayesian}) extend the LTH into the Bayesian framework by placing priors on weight-level mask variables and using variational inference to approximate the posterior over active weights. However, if \cite{frankle2019lottery} confirms that winning tickets generalise across architectures, \cite{liu2022winning} acknowledges that the procedure is expensive and provides no principled criterion for choosing the sparsity level. Moreover, this method was performed only at the weights level, not at the node level, and removing a neuron eliminates an entire learned feature representation.

     In Bayesian neural network, exact posterior inference in Bayesian neural networks is intractable. However, two broad families of approximation have emerged to handle this. Markov Chain Monte Carlo (MCMC) methods, pioneered by Neal (\cite{neal1996bayesian}), draw samples from the true posterior and are asymptotically exact, but their computational cost scales poorly with network size, making them impractical for modern architectures. Variational Inference (VI) offers a more scalable alternative: it reframes posterior computation as an optimisation problem, searching within a tractable family of distributions for the member closest to the true posterior in KL divergence. Among variational families, the Mean-Field family is the most widely adopted (\cite{xing2012generalized}).
     
     Bayesian neural networks (BNNs) place prior distributions over weights and use Bayes' theorem to compute posterior weight distributions, providing uncertainty quantification absent from standard networks. Neal (\cite{neal1996bayesian}) established the theoretical foundation but noted that exact posterior inference via MCMC is intractable for networks of practical size. Modern BNNs circumvent this through variational inference: Blundell et al.\ (\cite{blundell2015weight}) introduced Bayes by Backprop, which minimises the KL divergence between a factorised Gaussian variational family and the true posterior. The practical accessibility of BNNs was further broadened by Gal and Ghahramani (\cite{gal2016dropout}), who proved that dropout training (already standard practice) is mathematically equivalent to approximate Bayesian inference in a deep Gaussian process.
	
	A limitation unites all these approaches: they use continuous weight priors. Consequently, the posterior probability that any specific weight is exactly zero is zero. Continuous priors cannot provide genuine sparsity. Molchanov et al.\ (\cite{molchanov2017variational}) made partial progress by showing that variational dropout with an unbounded dropout rate naturally shrinks some weights toward zero, achieving effective sparsity. However, the resulting zeros are approximate, and there is no principled mapping from dropout rates to posterior inclusion probabilities for individual neurons.

     The above challenges were also stated by~\cite{mlambo2024survey}, who highlighted that mean-field variational methods don't scale to deep architectures. However, \cite{jantre2022layer} developed a method like SS-Independent Gaussian using a mean-field variational family, applied at the neuron level, with demonstrated performance and consistency. But no existing method uses Atom Prevalence as the generative framework for neural feature selection.
	
    This paper proposes a framework that bridges the gap identified at the end of the previous paragraph. It treats each neuron as an atom in the sense of \cite{fokoue2008estimation}, assigns it a posterior inclusion probability through a spike-and-slab variational posterior, and selects the final architecture by identifying the most likely layer size and retaining the most prevalent neurons. Concretely, the spike-and-slab Independent Gaussian prior of \cite{jantre2022layer} provides the variational machinery for computing neuron-level inclusion probabilities within a feedforward network, while the Bayesian Lottery Ticket initialises the procedure from a compressed, well-conditioned subnetwork rather than a redundant dense one. The Poisson-Binomial distribution then replaces the fixed cut-off of the Median Probability model with a data-driven optimal layer size, mirroring the adaptive threshold that Fokou\'e introduced for linear atoms. The result is a framework that simultaneously inherits the exact sparsity of spike-and-slab priors, the compression efficiency of iterative magnitude pruning, and the probabilistic selection logic of Atom Prevalence, extended for the first time to the nonlinear, layered representations of deep networks.

\section{Proposed Method: Neural Atom Prevalence}
\label{sec:methodology}

The proposed method can be categorized as neuron model selection procedure in Bayesian Neural Network, based on the selection of neurons to form a BNN. The principle or skeleton is similar to the Atom prevalence of Fokou\'e (\cite{fokoue2008estimation}) but applied in deep learning and with different ingredients. The core idea of the method is to compress or reduce the size of a large network architecture while preserving the stability, the interpretability, and the performance in terms of accuracy.  The entire body of this chapter is dedicated to the construction of the method, covering the problem setup and notation, the theoretical development of the neural atom prevalence method, the posterior approximation, the neural atom selection rule, and finally, the neural atom prevalence method pipeline and the algorithm. The structure of the chapter will follow the order of the above steps.

\subsection{Problem setup and notation}
Consider the machine learning task consisting of building a model that explains a relationship between an output variable $Y$ and a feature vector $X=(x_1,...x_d)$ ($d$ is the feature dimension) using a data set $D = \{(X_1, Y_1), \dots, (X_n, Y_n)\}$($n$ is the dataset size), observations which are drawn from the joint distribution $P_{XY}$ of $(X, Y)$. This means finding the best approximation function $\eta_{\theta}$(a function depending on parameters $\theta$) of the true function $\eta_{0}$ mapping the feature $X$ to the output $Y$. In this work, the general form of the true model (true functions) is defined as follows:
\begin{equation}
    Y_i = \eta_o(X_i) + \epsilon_i, \quad i = 1, \dots, n 
\end{equation}

where $\epsilon_i \sim \mathcal{N}(0,\sigma_e^2)$ and $\eta_o(\cdot): \mathbb{R}^d \longrightarrow \mathbb{R}$.
With this configuration, the conditional distribution of $Y$ given $X$ under this true model is (\cite{jantre2022layer}):
\[ f_o(Y|X=x) = \frac{1}{\sqrt{2\pi\sigma_e^2}} \exp\left(-\frac{(y-\eta_o(x))^2}{2\sigma_e^2}\right) \]

Since we are working in the neural networks space, this work aims to construct the neural network model $\eta_{\theta}$ that gives the best approximation $f_{\theta}$ of the function $f_0$ where:
\begin{equation*}
     f_{\theta}(Y|X=x) = \frac{1}{\sqrt{2\pi\sigma_e^2}} \exp\left(-\frac{(y-\eta_{\theta}(x))^2}{2\sigma_e^2}\right)
\end{equation*}
and which is sparse, confident and interpretable. 

With the assumption of independence between the observations, the likelihood of the data under the model is:
\begin{equation}
 P_{\theta}^n = \prod_{i=1}^n f_{\theta}(y_i|x_i)   
\end{equation}

For high computational issues, as suggested by Jantre et al. (\cite{jantre2022layer}), the following assumptions will be considered:
$P_X = U[0,1]$, $\sigma_e^2 = 1$, and that all activation functions $\psi_l$ in the work are 1-Lipschitz continuous.

\subsubsection{Neural Network architecture}
A large part of research on neural networks has focused on the weights, considering them as the basic unit for the composition of these architectures, and this is probably due to the importance or the place of the weights for traditional machine learning, like linear regression.  In the field of model size reduction, a lot of work has been carried out at the weight level, achieving important and very useful results. However, the price of these achievements is heavy, especially in the field of deep learning, where the weight space dimension grows extremely faster as the network becomes deeper. Shifting from weight space to neuron space can consistently reduce the dimensionality of the problem (moving from dimension D = 27 in weight space to D=10 in the neuron space for a small-sized network). 
Since we are doing a research,  we will first construct the method for the basis neural architecture, the Feedforward Neural Network (FNN). In that case, all the chapters will then be based on how to select neurons in a large FNN to achieve good dimension reduction (or network compression),  model interpretability, level of stability and also a good accuracy. Let define mathematical the FFNN: 

let consider a FNN with $L$ hidden layers, each which $K_{l}$ ($l \in \{1,2,...,L\}$) neurons, an input layer with $K_{0}$ input variables ($ X = {x_{i}, i \in \{1,2,...,d\}}$) and finally an output layer with $K_{L+1}$ dependent variables (${y_{i}, i \in \{1,2,...,K_{L+1}\}}$). Let denotes by $W^{l}= (w^{l}_{ij}, 1\leq i \leq  \{1,2,...,K_{l}\}, j \in \{1,2,...,K_{l+1} \})$ for all $0 \leq l \leq L$ and $B^l = (b^l_1, ..., b^l_{K_{l+1}} )$, respectively the matrix of weights and the vector of bias connecting layer $l$ to layer $l+1$. Therefore the general form of the neural network function is given by the following equation:
\begin{equation*}
    \eta_{\theta}(x) = \psi_{L+1}(B^{L} + W^L \psi_{L}(B^{L-1} + W^{L-1} \psi_{L-1}(\dots(B^0 + W^0 x))))
\end{equation*}

where the parameters is $\theta= \{W^l, B^l\}_{l=0}^L$ and  $\psi_l$ ($l = 1, \dots, L+1$) the activation function in layer $l$. 
\subsubsection{Neural atoms}

The construction of $\eta_{\theta}$ while operating at the neuron or node level requires mathematically defining what a neuron is. Operating at the neuron level implies considering neurons as the basic components of the neural model, the neural atom. So, defining mathematically what a neuron is, is defining what a neural atom is.

\begin{definition}[Neural atom] 
    A neural atom or an activation unit is the function $h^{l}_{j}$ defined for all $1 \leq l \leq L$ \text{and} $1 \leq j\leq K_l$ by:
    \begin{equation}
       h^{l}_{j}( h^{l-1}, W^{l-1}_j, b_{l-1})=\begin{cases} 
      \psi_1(\sum^d_iw^{0}_{ij}x_{i} + b_0) & \text{if } l =1 \\
      \psi_{l}(g_{lj}) & \text{if } 2 \leq l \leq L 
   \end{cases}
    \end{equation}
    where:
    \begin{itemize}
        \item $h^{l-1} = (h^{l-1}_1,...,h^{l-1}_{K_{l-1}})$, the neural atoms of layer $l-1$ ;
        \item $W^{l-1}_j = (w^{l-1}_{1j},...,w^{l-1}_{K_{l-1}j})$ the vector of weights incident to neural atom $h^{l}_{j}$ so that $w^l_{ij}$ can be represented by $\left(h_i^{l-1},h_j^l\right)$ ;
        \item $\psi_l$ is an activation function;
        \item $g_{lj}$ is an preactivation unit define by:
         \begin{equation}
          g_{lj} = W^{l-1}_j\times h^{l-1}+b_{l-1} = \sum^{K_{l-1}}_iw^{l-1}_{ij}h^{l-1}_{i} + b_{l-1}
         \end{equation}
    \end{itemize}  
\end{definition}

With the above definition, it can be derived that the reduced model will be a function of some of the neural atoms, selected from the set $\mathcal{A} = \{h^{l}_{j}, \quad  1 \leq j \leq K_l, \quad 1\leq l \leq L  \}$ of all atoms. So the Neural Atom Prevalence method aims to select the most prevalent neural atoms to construct $\eta_{\theta}$. 

Before moving, let's do the following remark~\ref{rem:mark}:

\begin{remark}{(Layer non-collapse constraint)}
     \label{rem:mark}
A standard feedforward network with $L$ hidden layers calculates a function $f_\theta: \mathbb{R}^{K_0} \rightarrow \mathbb{R}^{K_{L+1}}$ by connecting $L + 1$ mathematical steps in a row. Each step changes the numbers and then uses an activation function. Therefore, there must exist at least one valid path of adjacent weights linking the input layer to the output layer. If an entire layer is deactivated during neuron selection, no information can propagate through the network. Consequently, the mapping defined by the neural network no longer exists. Hence, to guarantee the existence of a valid functional model, at least one neuron must remain active for each hidden layer.
\end{remark}
Since this research aims to develop a framework that starts from a dense neural network and selects a subset of neurons to form a smaller subnetwork, the remark~\ref{rem:mark} will guide us for the next section on the presentation of the NAP generative model.

\subsection{Theoretical development of the Neural Atom Prevalence Method: the NAP Generative Model}

This method is built in full Bayesian(for uncertainty quantification) upon the work of Jantre et al. (\cite{jantre2022layer}) and Fokou\'e~(\cite{fokoue2008estimation}), which introduced the spike-and-slab prior for model parameter selection in multi-linear regression models. It is improved here for node selection by assigning a group prior to all weights incoming to a node, as in \cite{jantre2022layer}. The rest of this section will present the full probabilistic model of $\eta_{\theta}$ before observing the data, which depends on two groups of parameters, $\theta$ and the neuron indicator $V$, which will be defined below.  

\subsubsection{Prior for Neural atom inclusion, model selection probability and model size distribution}

A priori from the original atom paper, atom inclusion follows a Bernoulli distribution of parameter $\pi_i$, independently from atom $j \neq i$. 
The naive use of this cannot work here since the probability of a neuron in layer $l$ is zero if all the neurons of the previous layer have been excluded. Therefore, this remark~\ref{rem:mark} justifies the choice of this work to borrow the concept of adaptive layer selection from~\cite{jantre2022layer}, by treating the hidden layer independently based on the fact that this layer will exist in the final model architecture. In this consideration, the number of possible models is then:
\[
N = \prod_{l=1}^{L} (2^{K_l} - 1)
\]

\subsubsection{model selection probability}
\label{subsec:priorNode}
For every neuron $h^l_j$ let denote by $v^l_j$ ($v^l_j \in \{0,1\})$ the variable indicating the selection status of $h^l_j$. As in \cite{fokoue2008estimation,jantre2022layer} let consider that apriori all $v^l_j$ independently follow Bernoulli distribution of parameter $\pi^l$ :

$$v^l_j \sim \text{Bernoulli}(\pi^l), \quad 1 \le j \le K_l  \quad 1 \leq l \leq L$$

Then, for each layer $l, \quad  1 \leq l \leq L$: 

  $$ V^l = (v_1^l, \dots, v_{K_l}^l) \sim \prod_{j=1}^{K_l} \text{Bernoulli} (\pi^l)$$ 
  
such that for a model $M_V$ represented by the vector $V=(V^1,...V^L)$ indicating for each layer which nodes are selected to be part of $\eta_{\theta}$, the prior $P(V)$ is given by:

\[P(V) =\prod_{k=1}^{L} P(V^l) \quad where \quad V^l\sim\prod_{i=1}^{K_l} \text{Bernoulli}(\pi^l)\]

This is the prior probability for a given model (subset of hidden neuron atoms) to be selected.
More explicitly, from the aforementioned theorem \ref{rem:mark} and the independence, we have that: 
\begin{align*}
    P(V^l)&=\prod^{K_l}_{j=1}(\pi^l)^{v^l_j}(1-\pi^l)^{1-{v^l_j}}\\
           &=(\pi^l)^{\sum^{K_l}_{j=1}{v^l_j}}(1-\pi^l)^{K_l-\sum^{K_l}_{j=1}{v^l_j}}
\end{align*}
such that the the detailed expression of $P(V)$ is given by:
\begin{equation}
\label{eq: priorV}
    P(V) = \prod^L_{l=1}(\pi^l)^{\sum^{K_l}_{j=1}{v^l_j}}(1-\pi^l)^{K_l-\sum^{K_l}_{j=1}{v^l_j}}
\end{equation}

In the next section, the parameter $\pi =(\pi^l,..,\pi^L)$ will be determined by the theoretical calibration formula derived in~\cite{jantre2022layer}. 

\subsubsection{Model Size Prior}
Given that the method proposed assumes that the input layer and output layer are fixed from the beginning (neural atoms are hidden nodes), then the model size here is essentially determined by the size of the hidden layers. So for a model $M_V$  it size is given by the vector $K_V =(K_{V^1},..., K_{V^L})$ where for $1 \leq l \leq L$, $K_{V^l} = \Vert V^l \Vert_0$  and 
$\Vert V^l \Vert_0$ is the number of non-zero components of $V^l$, the number of atoms in layer $l$.Under the prior defined in section~\ref{subsec:priorNode}, we then have that $K_{V^l} = \sum^{K_l}_{j=1}v^l_j$ is the sum of independent Bernoulli distributions with the same parameter $\pi^l$, which is clearly easy to determine due to the common parameter for all nodes in that layer. $K_{V^l}$ Consequently, follow a Binomial Distribution of parameter $(K_l, \pi^l)$, so  the explicit probability for a layer l to have size k is given by the following equation:
\begin{equation}
\label{eq:priorKv}
P(K_{V^l}=k) =\binom{K_l}{k} ({\pi^l})^k (1 - {\pi^l})^{K_l - k}
\end{equation}
This is the prior distribution of the layer size for each hidden layer in the network, and its posterior, which will be used by NAP for atom selection, is the Poisson Binomial distribution, where each atom $h^l_j$ has its own inclusion probability $\tilde{\pi}^l_j$, which will be defined in the following sections.

\subsubsection{Neural Atom Prior with Spike-and-Slab: Weights and Bias}
This section is largely based on the work of Jantre et al. (\cite{jantre2022layer}). In their work, the neural atom is treated via a latent variable-based prior and it prior distribution in this work is the grouped distribution of $\overline{W}_{lj}=(W^{l-1}_j, b_{l-1})=(w^{l-1}_{1j},...,w^{l-1}_{K_{l-1}j}, b_{l-1})$. This grouped distribution is the Spike and Slab distribution, where the spike is a Dirac spike, represented by $v^l_j$. The slab is the standard Gaussian (centered at $\mu_0=0$ with variance $\sigma^2_0=1$) distributed random variable. More formally, for an atom $h_j^l$, we have that:
\begin{equation}
\label{eq:priorW}
    \overline{W}_{lj} | v_j^l \sim (1 - v_j^l)\delta_0 + v_j^l \mathcal{N}(0, \sigma_0^2 I)
\end{equation}
where $\overline{W}_{lj}$ is the weight and bias vector that connects the layer $l-1$ to the node $j$ in layer $l$.
$\delta_0$ denotes the Dirac measure concentrated at zero, applied component-wise
to each element of $\overline{W}_{lj}$. From the independence of components of $\overline{W}_{lj}$ and the Gaussian Slab result, the SS-IG (Spike and Slab Independent Gaussian) prior for neural atoms. 

When $v_j^l=0$, the atom is inactive, and all its incoming weights and bias are forced to exactly zero by the Dirac spike $\delta_0$. When $v_j^l=1$, the neuron is active, and its weights are given a centred Gaussian slab with variance $\sigma_0^2$, encoding the prior belief that active weights are non-negligible but not systematically large.

Finally, under the independence  assumption across atoms, the full weight prior is: 
\begin{equation}
    P(\theta)= \sum_VP(\theta,V)=\sum_VP(\theta|V)P(V)=\sum_V\prod_{l=1}^L\prod^{K_l}_{j=1}P(\overline{W}_{lj} | v_j^l)P(v_j^l)
\end{equation}

And then for the full joint generative model, the full joint distribution of (X, Y) is:
\begin{equation}
    P_{XY}=\sum_{\theta,V}P(X,Y|\theta,V)=\sum_{\theta,V}P(Y|X,\theta,V)P(\theta|V)P(V)
\end{equation}

We finally fully specify the NAP generative model through the joint prior $P(\theta,V)$, which is composed by two main components, the $P(V)$, indicating the prior beliefs about sparsity, derived from independent Bernoulli and $P(\theta|V)$ the prior of weights. The next section of this work will focus on deriving the posterior $P(\theta,V|D)$ through Variational Inference due to its tractability compared to MCMC and also as recommended by~\cite{jantre2022layer}.

\subsection{Posterior approximation via Variational Inference}

Since the work is carried out at the node level, but driven by the weight distribution, the posterior of an activation unit $h_j^l$ is given by the marginal joint posterior of $(\overline{W}_{lj}, v_j^l)$. This is considered so that posterior of all activation units are driven by the posterior of all the weights, bias $\overline{W}$ and $V$, $P(\theta, V \mid D)$, where $\theta$ is the set of parameters and $V$ is the whole model indicator. This posterior has been derived by Jantre et al. \cite{jantre2022layer} and is:

\[
P(\theta, V \mid D) = \frac{P_\theta^n \; P(\theta \mid V) \; P(V)}
{\displaystyle\sum_{V} \int P_\theta^n \; P(\theta \mid V) \; P(V) \; d\theta}
\]

As is widely known in BNN, this posterior is not tractable, especially in deep architectures. To move forward with this, we will apply variational inference to choose a posterior distribution close to the true posterior and that is more tractable. The method is based on identifying a posterior family calling variational family, especially the Mean-Field-Family (\cite{xing2012generalized}), which will serve as a space for running an optimization algorithm to choose the variational distribution $q^*$ that minimize  $d_{KL} (q, P(.|D))$, which denotes the KL-distance between q and the true posterior $P(.|D)$.

\subsubsection{Variational Family Method for Posterior: Mean Field family.}

The variational family considered in this study is an adapted version of the Mean Field family of Jantre et al.  (\cite{jantre2022layer}), adjusted to take into account the variational posterior of the model size distribution. This is defined as follows:
\begin{equation}
   Q^{MF} = \left\{ \overline{W}_{lj} \mid v_j^l \overset{i.d}{\sim}
\left[ (1 - v_j^l)\, \delta_0 +
v_j^l \; \mathcal{N}\!\left(\mu_j^l,\, \mathrm{diag}\!\left((\sigma^l_{j})^{2}\right)\right) \right],\; K_V \sim \sum_{V \in \{V,\; \|v\|_0 = K_V\}} q(V), \;  v_j^l \overset{i.d}{\sim} \mathrm{Ber}(\tilde{\pi_{j}}^l)\right\} 
\end{equation}

where $\mu_j^l = \left(\mu_{1j}^l, \ldots, \mu_{K_{l-1}j}^l, \mu_{K_{l-1}+1j}^l\right)$
and  $\left(\sigma_j^l\right)^2 = \left\{ \left(\sigma_{1j}^l\right)^2, \ldots \left(\sigma_{K_{l-1}j}^l\right)^2, \left(\sigma_{K_{l-1}+1j}^l\right)^2 \right\}$
are respectively the vectors of variational mean and standard deviation parameters of weights and biases incident on node $j$ in layer $l$.

Given all this, the approximated posterior is given by:
\begin{align*}
q^*(\theta, V) &= \underset{q \in Q^{MF}}{\arg\min}
\sum_{V} \int \left[\log q(\theta, V) - \log P(\theta, V \mid D)\right] q(\theta, V)\, d\theta\\
&= \underset{q \in Q^{MF}}{\arg\min}
\left(
\sum_{V} \int \left[\log q(\theta, V) - \log P(\theta, V, D)\right] q(\theta, V)\, d\theta
+
\log\!\left(\sum_{V} \int P_\theta^n\, P(\theta \mid V) P(V)\, d\theta\right)
\right)\\
&= \underset{q \in Q^{MF}}{\arg\min}
\left[
- \mathrm{ELBO}(q,\, P(\cdot \mid D))
\right]
+ \log\!\left(\sum_{V} \int P_\theta^n\, P(\theta \mid V) P(V)\, d\theta\right)\\
&= \underset{q \in Q^{MF}}{\arg\max} \; \mathrm{ELBO}(q,\, P(\cdot \mid D))
\end{align*}
where
\begin{equation}
    \mathrm{ELBO}(q, P(\cdot \mid D))
= \sum_{V} \int \left[\log P(\theta, V, D) - \log(q(\theta, V))\right]
q(\theta, V)\, d\theta
\end{equation}

The resolution of the above optimization problem will produce the marginal variational posterior of $\theta$, $q^*$, such that the variational posterior of the activation units is $q^*(\theta \mid V)$.
Just as in Jantre et al.\ (2022) let us denote by $Q^*$ the probability distribution function corresponding to $q^*$, where:
\begin{equation}
\label{eq: varpost}
    Q^*(A) = \int_A q^*(\theta)\, d\theta 
          = \sum_{V} \int_A q^*(\theta \mid V) P(V)\, d\theta
\end{equation}

\subsubsection{Consistency of the method}
This probability distribution
function is consistent. In fact, by allowing different numbers of activation units and different weight norm magnitudes per layer in the network, and by introducing a more flexible sparsity parameter (different regarding the layer), Jantre et al.\ provide a theoretical proof of consistency of the variational probability distribution $Q^*$. More especially, they show that $Q^*$ converges to the true posterior $P(\cdot \mid D)$ as $n$ (data size) goes to infinity under the following keys conditions established in \cite{jantre2022layer}:
\begin{itemize}
    \item the activation function must be $1$-Lipschitz continuous (satisfied by swish, ReLU, tanh, and sigmoid);
    \item network width and weight magnitudes are controlled and governed respectively by layer-dependent upper bounds $S_l=K_l$ rather than a uniform node count, accommodating the non-uniform topologies that arise after pruning and by per-layer $\ell_1$ bounds $B_l$ on incident weights rather than a single global bound, permitting sparse layers with heavier surviving weights;
    \item prior inclusion probabilities $\pi^l$ are specified layer-wise, with $\pi^{L+1} = 1$ fixed
at the output, ensuring optimal node recovery per layer and valid contraction rates.
\end{itemize} 
Together, these conditions guarantee that $Q^*$ contracts around the true regression function
$\eta_0$ at the minimax-optimal rate as $n \to \infty$, directly justifying the consistency claim
for the NAP sparse architecture whose topology varies across the IMP, SS-IG, and fine-tune phases.

With this notation (\ref{eq: varpost}), we have that:
	\begin{align*}
		\text{ELBO}(q,\, P(\cdot|D))
		&= \sum_V -\int \left[\log q(\theta, V) - \log P(\theta, V, D)\right] q(\theta, V)\, d\theta \\
		&= -\sum_V \int \log[q(\theta,V)]\, q(\theta,V)\, d\theta + \sum_V \int \log(P(\theta,V,D))\, q(\theta,V)\, d\theta \\
		&= -\sum_V \int \log(q(\theta,V))\, q(\theta,V)\, d\theta + \sum_V \int \log(P(D|\theta,V)\, P(\theta,V)) q(\theta,V)\, d\theta \\
		&= -\sum_V \int \log(q(\theta,V))\, q(\theta,V)\, d\theta + \sum_V \int \log\!\left(P_\theta^n\right) q(\theta,V)\, d\theta + \sum_V \int \log(P(\theta,V))\, q(\theta,V)\, d\theta \\
        &=\sum_V \int \log\!\left(P_\theta^n\right) q(\theta,V)\, d\theta - \left[-\sum_V \int \log(P(\theta,V))\, q(\theta,V)\, d\theta + \sum_V \int \log(q(\theta,V))\, q(\theta,V)\, d\theta \right]\\
        &=\sum_V \int \log\!\left(P_\theta^n\right) q(\theta,V)\, d\theta - \sum_V \int \left[\log(q(\theta,V))-\log(P(\theta,V))\right]\, q(\theta,V)\, d\theta \\
		&= \mathbb{E}_q\!\left(\log(P_\theta^n)\right) - \sum_V \int q(\theta,V)\log\!\left(\frac{q(\theta,V)}{P(\theta,V)}\right) d\theta \\
		&= \mathbb{E}_q\!\left[\log P_\theta^n\right] - d_{KL}(q\,,P) \\
		&= \mathbb{E}_{q(\theta|V)q(V)}\!\left[\log P_\theta^n\right] - d_{KL}\!\left(q(\theta|V)q(V) \,\|\, P(\theta|V)P(V)\right) \quad\text{Over all possible $V$}
	\end{align*}
	
	From chain rule of $d_{KL}$:
	\[
	\text{ELBO}(q,\, P(\cdot|D))= \underbrace{\mathbb{E}_{q(\theta|V)q(V)}\!\left[\log P_\theta^n\right]}_{a}
	- \left[\underbrace{d_{KL}(q(V)\|P(V))}_{b}
	+ \underbrace{\mathbb{E}_{q(V)}\!\left[d_{KL}(q(\theta|V)\|P(\theta|V))\right]}_{c}\right]
	\]
	
	Let's continue the development by extending first Components $b$ and $c$ of ELBO. This will let us recall the following formulas: Starting with the variational distribution, we have
	
	\[
	q(\theta|V) = \prod_{l=1}^{L} q(W^l, B^l | V^l)
	= \prod_{l=1}^{L} \prod_{j=1}^{K_l} q(\overline{W}_{lj} \,|\, {v}_j^l)
	\]
	\[
	q(V) = \prod_{l=1}^{L} \prod_{j=1}^{K_l} \text{Bernoulli}\!\left(\tilde{\pi}_j^l\right)
	\]
	
	And $P(\theta|V)$ and $P(V)$ are defined analogously and refer to prior distribution, with $\theta|V$ distributed as in \ref{eq:priorW}, and $V$ as in \ref{eq: priorV}, based on the same prior inclusion probability per layer.
	
	We can now extend the Component $b$ as follows:
	\begin{align}
		b &= d_{KL}(q(V)\|P(V))\\
		&= \sum_{lj} \int q(v_j^l)\log\!\left[\frac{q(v_j^l)}{P(v_j^l)}\right] dv_j^l\\
		&= \sum_{lj} d_{KL}(q(v_j^l)\|P(v_j^l))\\
		&= \sum_{lj}\left[\tilde{\pi}_j^l \log\frac{\tilde{\pi}_j^l}{\pi^l} + (1-\tilde{\pi}_j^l)\log\frac{(1-\tilde{\pi}_j^l)}{(1-\pi^l)}\right]
	\end{align}
	
	Which is the sum of the KL distance between the variational distribution $q$ and the prior $P$ for every inclusion indicator variable for all the neural atoms. This quantity is differentiable with respect to $\tilde{\pi}_j^l$, which is important for training.
	
	Concerning the Component $c$ we have:
	\begin{align}
		c &= \mathbb{E}_{q(V)}\!\left[d_{KL}(q(\theta|V)\|P(\theta|V))\right]\\
		&= \mathbb{E}_{q(V)}\!\left[\sum_{lj} d_{KL}\!\left(q(\overline{W}_{lj}\,|\,v_j^l) \,\middle|\, P(\overline{W}_{lj}\,|\,v_j^l)\right)\right]\\
		&= \sum_{lj} \mathbb{E}_{q(v_j^l)}\!\left[d_{KL}\!\left(q(\overline{W}_{lj}\,|\,v_j^l) \,\middle|\, P(\overline{W}_{lj}\,|\,v_j^l)\right)\right]
	\end{align}
	
	Since $v_j^l \in \{0,1\}$ we have:
	
	\begin{align*}
		c = \sum_{lj} q(v_j^l=1)\, d_{KL}\!\left(q(\overline{W}_{lj}\,|\,v_j^l=1) \,\middle|\, P(\overline{W}_{lj}\,|\,v_j^l=1)\right) &+ \\
		\sum_{lj} q(v_j^l=0)\, d_{KL}\!\left(q(\overline{W}_{lj}\,|\,v_j^l=0) \,\middle|\, P(\overline{W}_{lj}\,|\,v_j^l=0)\right)
	\end{align*}
	
	Since $q(v_j^l=0) = 1 - \tilde{\pi}_j^l$ and that under $v_j^l = 0$

	$q(\overline{W}_{lj}\,|\,v_j^l=0) = \delta_0 = P(\overline{W}_{lj}\,|\,v_j^l=0).$

	then
	\[
	d_{KL}\!\left(q(\overline{W}_{lj}\,|\,v_j^l=0) \,\middle|\, P(\overline{W}_{lj}\,|\,v_j^l=0)\right) = 0
	\]
	
	So
	\[
	c = \sum_{lj} \tilde{\pi}_j^l\, d_{KL}\!\left(q(\overline{W}_{lj}\,|\,v_j^l=1) \,\middle|\, P(\overline{W}_{lj}\,|\,v_j^l=1)\right)
	\]
	
	\[
	= \sum_{lj} \tilde{\pi}_j^l\, d_{KL}\!\left(\mathcal{N}(\mu_j^l,\, \text{diag}(\sigma_j^{l2})) \,\middle|\, \mathcal{N}(0,\, \sigma_0^2 I)\right)
	\]
	
	This result from the variational family and prior of $\theta$. This quantity is also differentiable regarding $\mu_j^l$ and $\sigma_j^l$.
	
	If Components $b$ and $c$ of ELBO are differentiable regarding the parameters, it's not the case of the data-fit parameter Component $a$. Authors have noticed that and Jantre et al as well. This challenge is due to the discrete character of a $v_j^l$. As with previous research (Jang et al 2017, Maddison et al 2017, \cite{jantre2022layer, jang2017categorical, maddison2017concrete}) we will consider the Gumbel-Softmax approximation that approximate the distribution $q(v_j^l) \sim \text{Bernoulli}(\tilde{\pi}_j^l)$ by the Gumbel-Softmax distribution: $q(\tilde{v}_j^l) \sim \text{GS}(\tilde{\pi}_j^l, \tau)$. The appendix~\ref{app:additionalData} provides more details about this approximation, showing how it works practically and also provides weights posterior hyperparameter specification to complete the SS-IG frame.

\subsection{The Neural Atom Prevalence selection rule}
In the previous section we specify step by step how to compute the variational posterior distribution of weights, bias and also the neural atom inclusion posterior probability $\tilde{\pi_j}^l$ for each neural atom $h^l_j$. This inclusion posterior tell how likely the atom is to be selected for the model $\eta_{\theta}$.  An this last posterior depend on the prior inclusion per layer $\pi^l$, but we haven yet specify hoe it's obtain.

\subsubsection{$\pi^l$ calibration from the Bayesian Lottery Ticket}
To complete the method of Jantre et al. (\cite{jantre2022layer}) we have to specify how to obtain or choose $\pi^l$ and $\tau$.
	
Let's start with $\tau$ whose the most suitable value, serving as trade off between discrete $v_j^l$ and the continuous approximate $\tilde{v}_j^l$, is $\tau = 0.5$.
	
For $\pi^l$, the paper by Jantre et al. (\cite{jantre2022layer}), through it Corollary 4.5, achieves the proofs of the consistency of the method and the resulting variation posteriors, stating that the optimal value of $\pi_l$ is giving by the following equation \ref{eq:priorlambda}:
	
	\begin{equation}
	    \label{eq:priorlambda}
        -\log(\pi^l) = \log(k_{l+1}) + C_l(k_l+1)\mathcal{V}_l 
	\end{equation}
	\text{ where}
	\[
	\mathcal{V}_l = B_l^2/(k_l+1) + \sum_{\substack{m=0\\m\neq l}}^{L} \log(B_m) + L + \log(k_l+1) + \log(n) + \log\!\left(\sum_{m=0}^{L} \mathcal{U}_m\right)
	\]
	\[
	\mathcal{U}_m = (L+1)^2\!\left[\log n + \log(L+1) + \log(k_l+1) + \log(k_{l+1})\right]
	\]
    \[
	 \|(\|\overline{W}_{l1}\|_1,...,\|\overline{W}_{l{K_l}}\|_1)\|_\infty\leq B_l 
	\]
where the $B_l= K_{l-1}+1$ in~\cite{jantre2022layer} they select $C_l$ in the order of $10^{-5}$ to avoid having $\pi^l<10^{-50}$.  This work take a different approach to have a more realistic $\pi^l$.  The Bayesian Lottery ticket(the winning ticket) is a good departure for improving the model sparsity while preserving prediction preformance. According to~\cite{kuhn2026bayesian} BLT can prune up to 80\% of model weights without harming the prediction performance. To take advantage of this fact, the NAP method compute  $B_l=B_{wt}$ recorded on the winning ticket and $C_l$ is calculated such that  $10^{-50}\leq\pi^l\leq S_{wt}$ were  $S_{wt}$ is given as follow:
\[S_{wt}=\|(\|\overline{W}_{l1}\|_1,...,\|\overline{W}_{l{K_l}}\|_1)\|_0\]
calculated on the winning ticket.

\subsubsection{Posterior distribution of the Layer Size $K_{V^l}$:} 
		The first motivation of the adjustment above is to obtain the inclusion posterior probability of $\tilde{\pi}_{j}^{l}$ for each neural atom.
		
		Recalling that for each layer $l$:
		$$K_{V^l}=\sum_{j=1}^{K_l}v_{j}^{l} \quad (v_j^l \sim \text{Bernoulli} (\tilde{\pi}_{j}^{l}))$$
		
		The variable $K_{V^l}$ follows a Poisson-binomial distribution (\cite{hong2013computing}):
		$$P(K_{V^l}=k)=\sum_{\substack{V^l\\\|V^l\|_0=k}}\prod_{j=1}^{K_l} ({\tilde{\pi}_{j}^{l}})^{v_j^l} (1-\tilde{\pi}_j^{l})^{1-v_j^l}$$
		
		but for very dense networks this distribution is intractable (\cite{hong2013computing}). Many algorithm has been develop to overcome this issue among which the Dynamic Programming Recursion (\cite{hong2013computing})  with which is give as follow:
		
		For $k$:
        Let $K_{V^l}^{(j)}$  denote the number of active neurons among the first $j$ neurons in layer $l$, with the convention $K_{V^l}^{(0)} = 0$. The distribution of the total active node count $K_{V^l} = K_{V^l}^{(K_l)}$ is then computed via the recursion
        
        \[
        P\!\left(K_{V^l}^{(j)} = k\right)
        = P\!\left(K_{V^l}^{(j-1)} = k\right)\left(1 - \tilde{\pi}_j^l\right)
        + P\!\left(K_{V^l}^{(j-1)} = k - 1\right)\tilde{\pi}_j^l,
        \]
        
        initialised at $P(K_{V^l}^{(0)} = 0) = 1$ and $P(K_{V^l}^{(0)} = k) = 0$ for $k > 0$.
        This forward recursion has complexity $\mathcal{O}(K_{\max}^2)$ and is considerably
        more tractable than enumerating all $2^{K_l}$ node subsets directly.
		
		\subsubsection{ The most likely layer size  $K_{l}^{op}$}
		With the posterior distribution of the layer $l$, the most likely layer size of the distribution is given by:
		$$K_{l}^{op}=\text{arg max}_k P(K_{V^l}=k)$$
		
		\subsubsection{Selection of the most prevalent Neural atom}
		The selection is performed by simply rank the atom in each layer according to their inclusion posterior $\tilde{\pi}_{j}^{l}$
		
		Then Atom $h_{j}^{l}$ is selected if 
		$$\mathcal{R}(\tilde{\pi}_{j}^{l}) \le K_{l}^{op}$$
		
		where $\mathcal{R}(\tilde{\pi}_{j}^{l})$ is the rank of $\tilde{\pi}_{j}^{l}$ in the sorted list of $\tilde{\pi}_{j}^{l}$ in layer $l$.
        
	\subsection{Neural Atom Prevalence method pipeline and Algorithm}
	
	The final NAP method combines successively the BLT through Iterative Magnitude Pruning and, the Soft SS-IG and PB selection and then the fine-tuning of the selected network. To finalise the pipeline, the following adjustments need to be made from the original SS-IG by~\cite{jantre2022layer}.
  Avoid pruning during training: To achieve this, the soft variable $\tilde{v}_j^l$ will be used in both the forward and backward pass during training. This will keep all the neurons until the end of the training, allowing to compute both the inclusion posterior and the model size distribution. That is why we are talking about soft SS-IG on winning ticket.
	
\subsubsection{Method Algorithm: Neural atom prevalence method}
The following table~\ref{tab:nap_algorithm} provides a detailed algorithm of the Neural Atom Prevalence pipeline.
    \begin{table}[H]
    \centering
    \caption{Neural Atom Prevalence (NAP) Algorithm Structure}
    \label{tab:nap_algorithm}
    \begin{tabular}{|l|}
    \hline
    Algorithm: Neural Atom Prevalence (NAP) \\ \hline
    Input: Training data D, dense architecture, optimizer, hyperparameters \\ 
    Output: Sparse BNN with selected neurons \\ \hline
    \rule{0pt}{3ex}%
    1. Train Bayesian Lottery Ticket (BLT) on dense network \\ 
    2. Extract winning ticket weights $W^l$ and biases $B^l$ \\ 
    3. Compute layer-wise prior inclusion probabilities $\pi^l$ using Corollary 4.5 of Jantre et al. \\ 
    \hspace{3mm} substituting dense network statistics with winning ticket statistics ($S_{wt}$, $B_{wt}$) \\ 
    4. Initialize SS-IG on winning ticket architecture with soft $v_j^l$ in forward/backward passes \\ 
    5. Train SS-IG for $E$ epochs, tracking posterior inclusion probabilities $\tilde{\pi}_j^l$ \\ 
    6. For each layer $l$, compute layer size distribution $P(K_{V^l} = k)$ via Poisson-binomial DP recursion \\ 
    7. Find optimal layer size $K_l^{op} = \text{argmax}_k P(K_{V^l} = k)$ \\ 
    8. In each layer, select the $K_l^{op}$ neurons with highest $\tilde{\pi}_j^l$ \\ 
    9. Prune unselected neurons and their incident weights \\ 
    10. Fine-tune the sparse BNN from epoch-0 initialization of the original SS-IG \\ 
    11. Return final sparse network \\[1ex] \hline
    \end{tabular}
    \end{table}

We proposed a neural atom prevalence method for model selection, based on the prevalence of neural atom. Our method is an adjustment of the spike and slab independent gaussian method of \cite{jantre2022layer}. We modified the training process of their method and the determination of the model prior inclusion distribution probability using Bayesian winning ticket network architecture. This change allows us to transpose the Fokoue's atom prevalence method to neural network by computing the posterior inclusion probability for all the neural atom, the a posteriory most likely layer size for each layer in the network and then construct the final model by selecting the most prevalent atoms. this model as shown to be consistent, producing a model, that, converge to the true model as the dataset become larger.  The next chapter of this work will consist of experimenting the method with the aims of proving the quality of the model and empirically comparing it with the baseline SS-IG and the dense VBNN.

\section{Empirical Validation of Neural Atom Prevalence Method}
\label{sec:experiments}
This chapter serves as empirical validation of the method and algorithm developed in the previous chapter. Our framework is designed to be applicable to both regression and classification problems. To ensure scientific rigor, experiments are conducted on both real-world and simulated datasets. This is important since simulated data provide a known ground truth against which results can be exactly assessed. Two other models will be used in comparative analysis, the original SS-IG by Jantre et al. \cite{jantre2022layer} (baseline) and the dense VBNN as described in \cite{jantre2022layer}. The chapter is organized as follows: The first part will focus on the setup of the experiment, presenting the experiment environment from the selection of the data set to the specification of hyperparameters for all types of data (simulated, regression and classification). The second and third parts will present the results and discussions, showing the performance of the proposed method compared to the baseline and the winning ticket.

\subsection{Experimental Setup}
All experiments are conducted on a local Jupyter notebook equipped with an NVIDIA 6 GB VRAM GPU, running with Python 3.11 and CUDA. Random seeds are fixed at 42 across all libraries (PyTorch, NumPy) to ensure reproducibility. The Adam optimizer is used throughout with gradient clipping at norm 10.0. Three methods are compared in all experiments:

\begin{itemize}
    \item The proposed Neural Atom Prevalence Method: Applying SS-IG with soft~$\tilde{v_j}^l$ in both forward and backward passes on the winning Ticket network with weights. This is followed by the Poisson Binomial dynamic programming algorithm, applied to compute the optimal network size, which allows node selection to be performed. Finally, the selected network is then fine-tune in a Bayesian framework from the epoch\-0 of SS-IG initialization;
    \item The original SS-IG (by Jantre et al. 2022): The original Spike and Slab Independent Gaussian Bayesian neural network, trained on the full dense architecture using a hard Gumbel-Softmax relaxation (straight-through estimator) with layer-wise prior inclusion probabilities from Corollary 4.5 of \cite{jantre2022layer};
    \item The VBNN as presented in (\cite{jantre2022layer}). 
\end{itemize}

At the end of all experiments, the sparsity is reported as the fraction of nodes removed relative to the original dense architecture in each layer. All accuracy and Root Mean Squared Error (RMSE) values are reported on training, validation, and test sets across training epochs.

\subsubsection{Simulated Data}
The simulation study follows the design of Jantre et al. (2022, section 6.2) to allow for direct comparison. Data are generated from the nonlinear regression model:
$$Y = \frac{7x_2}{1+x_1^2}+\sin(x_3 x_4)+2x_5 + \varepsilon, \quad \varepsilon \sim \mathcal{N}(0,1)$$

Where all covariates $x_1, \ldots, x_5$ are i.i.d. $\mathcal{N}(0,1)$ and independent of $\varepsilon$. A training set of 3000 observations and a test set of 1000 observations are generated. Regarding the network architecture, a two-hidden-layer network with 20 neurons per layer is used, exactly as in~\cite{jantre2022layer}. Swish activation is used for all three methods for consistency with \cite{jantre2022layer}. The model is trained for 10000 epochs with a learning rate of $5\times 10^{-3}$ and full batch gradient descent. The evaluation metric is Root Mean Squared Error (RMSE) on the test set, along with layer-wise node sparsity.

The hyperparameters for the simulated data experiment are summarized in Table 4.1.
	
	\begin{table}[h!]
		\centering
		\caption{Simulated Data Hyperparameters}
		\label{tab:simulated}
		\begin{tabular}{lc}
			\toprule
			Parameter & Value \\
			\midrule
			Training samples & 3,000 \\
			Test samples & 1,000 \\
			Architecture & \( 5 \rightarrow 20 \rightarrow 20 \rightarrow 1 \) \\
			Activation & Swish \\
			Optimizer & Adam \\
			\bottomrule
		\end{tabular}
	\end{table}

\subsection{UCI Regression Datasets}
	Two real-world regression datasets from the UCI Machine Learning Repository ~\cite{dua2017uci} are used, following the experimental protocol in \cite{jantre2022layer}. Each dataset is split randomly into a 90\% training set and a 10\% test set. For the smaller dataset (Concrete), this procedure is repeated over 20 independent random splits, and results are reported as the mean \(\pm\) standard deviation of test RMSE. For the Year dataset, one split is used, following the original protocol.
	
	A single hidden layer network with 50 neurons is used for Concrete, and 100 neurons for Year, with Swish activation throughout. All models are trained with the Adam optimizer at learning rate \(10^{-3}\) and batch size 128 (256 for Year).
	
	Table~\ref{tab:uci} summarizes the UCI regression datasets and their respective configurations.
	
	\begin{table}[h!]
		\centering
		\caption{UCI Regression Dataset Summary}
		\label{tab:uci}
		\begin{tabular}{lcccccc}
			\toprule
            Dataset & \(\bm{n}\) & \(\bm{d}\) & Hidden units & Splits & Epochs & Batch \\
			\midrule
            Concrete & 1,030 & 8 & 50 & 20 & 500 & 128 \\
			Year & 515,345 & 90 & 100 & 1 & 100 & 256 \\
			\bottomrule
		\end{tabular}
	\end{table}
	
	\subsubsection{Prior Calibration and Pruning}
	The layer-wise prior inclusion probabilities \(\pi^1\) for SS-IG and NAP are computed from Corollary 4.5 of \cite{jantre2022layer}. For NAP specifically, \(\pi^1\) is computed using the surviving node count $S_{wt}$ and the maximum L1 weight norm $B_{wt}$ of the winning ticket, rather than the full dense architecture, calibrating the prior to the compressed topology. For the SS-IG baseline, $\pi^1$ is computed using the full dense architecture dimensions and \(B_1 = k_1 + 1\) as recommended in Appendix B.4 of \cite{jantre2022layer}.
	
	The IMP phase uses 5 pruning rounds with a pruning fraction of 0.35 per round. The evaluation metric is test RMSE in original (unscaled) units, obtained by inverting the \texttt{StandardScaler} applied during preprocessing.
	
	\subsubsection{Image Classification Datasets}
	The same standard image classification benchmark is used as in \cite{jantre2022layer}: MNIST. This dataset consists of 60,000 training images and 10,000 test images of size \(28 \times 28\) pixels across 10 classes. Pixel values are divided by 126, following the exact protocol of \cite{jantre2022layer}. Images are flattened to vectors of dimension 784 for the MLP architecture.
	
	\subsubsection{Architecture}
	A two-hidden-layer MLP with 400 neurons per layer and 10 output neurons is used (\(784 \rightarrow 400 \rightarrow 400 \rightarrow 10\)). The Swish activation function (\(f(x) = x \cdot \sigma(x)\), also known as SiLU) is used for all three methods, as recommended by the original paper to avoid the dying neuron problem in large networks (\cite{jantre2022layer}).
	
	\subsubsection{Training Configuration}
	All methods are trained for 1,200 epochs with batch size 1,024 and learning rate \(10^{-3}\), matching the paper's classification setup. The KL term is annealed linearly from 0 to its maximum value over the first 400 epochs (one third of training), following the original implementation of the original SS-IG. For the IMP phase, 5 pruning rounds are used with a pruning fraction of 0.35 per round, and 500 epochs of training per round. The fine-tuning phase of NAP runs for an additional 1,200 epochs on the pruned BNN topology.
	
	\subsubsection{Hyperparameters}
	Table~\ref{tab:class} lists the hyperparameters for the classification experiments.
	
	\begin{table}[h!]
		\centering
		\caption{Classification Experiment Hyperparameters}
		\label{tab:class}
		\begin{tabular}{lccc}
			\toprule
			Parameter & SS-IG & Winning Ticket & NAP \\
			\midrule
			Architecture & \(784 \rightarrow 400 \rightarrow 400 \rightarrow 10\) & \(784 \rightarrow S_1 \rightarrow S_2 \rightarrow 10\) & \(784 \rightarrow K^{op}_1 \rightarrow K^{op}_2\rightarrow 10\) \\
			Activation & Swish & Swish & Swish \\
			Epochs & 1,200 & 500 (final) & 1,200 (fine-tune) \\
			Batch size & 1,024 & 1,024 & 1,024 \\
			KL annealing & 0 \(\rightarrow\) 1 over 400 ep & -- & -- \\
			Prior \(\pi^l\) & \([\pi^1=0.005, \pi^2=0.003]\) & -- & Corollary 4.5 (ticket) \\
			IMP rounds & -- & 5 & 5 \\
			IMP prune fraction & -- & 0.35 & 0.35 \\
			\(\sigma^2\) & 1.0 & -- & 1.0 \\
			Temperature \(\tau\) & 0.5 & -- & 0.5 \\
			MC samples (test) & 10 & -- & 10 \\
			\bottomrule
		\end{tabular}
	\end{table}
	\noindent Here \(S_l\) denotes the number of surviving nodes in layer $l$ after IMP and \(K^{op}_l\) denotes the Poisson-Binomial optimal node count selected by NAP in layer $l$.
	
	\subsubsection{Metrics}
	Test accuracy is the primary performance metric. Node sparsity per layer is computed as the fraction of nodes removed relative to the original dense architecture (400 nodes per layer). For NAP, sparsity is fixed after PB selection and remains constant during fine-tuning. The compression ratio is defined as the ratio of active parameters in the compressed network to total parameters in the original dense network.
	
	\subsubsection{Weight Initialization Strategy}
	A key design choice of the proposed method, directly analogous to the Lottery Ticket Hypothesis, concerns the initialization of the final pruned BNN. In the LTH, after IMP identifies the winning ticket topology, the surviving weights are rewound to their epoch-0 values (the original random initialization of the dense network) before final training. The NAP method adopts the same principle for the Bayesian setting: after PB selection identifies the optimal \(K^{op}\) node topology from the trained SS-IG posteriors, the weights of the final BNN are initialized from the epoch-0 state of the original SS-IG that is, the Kaiming uniform initialization of \(\mu\), the constant initialization of \(\rho = -6\) (giving \(\sigma = e^{\rho/2} = e^{-3} \approx 0.05\)), and the prior inclusion probability \(\pi = 0.99\), rendering all nodes active. Only the node positions identified by PB are retained; all other nodes are discarded. This rewinding strategy ensures that the fine-tuning phase benefits from a clean, unbiased starting point calibrated to the selected topology, rather than inheriting potentially over-fitted posteriors from the full training run.

\subsection{Results and Discussion}
\subsubsection{Simulated Data experiment}

\begin{figure}[H]
    \centering
    \includegraphics[width=1.0\linewidth]{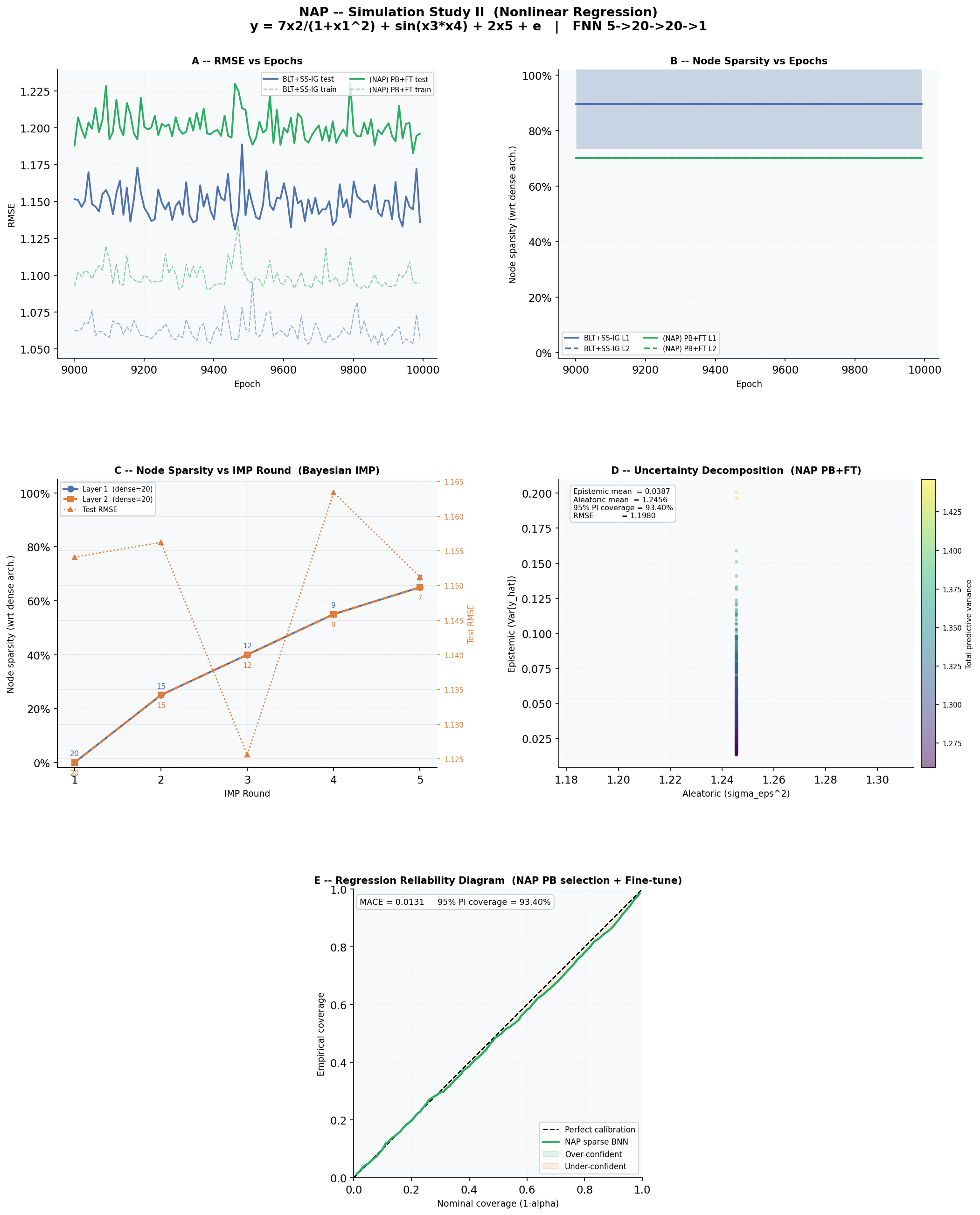}
    \caption{Results of experiment on simulated data:(A) RMSE vs epoch, NAP vs baseline. (B) Node sparsity vs epoch. (C) Node sparsity vs IMP round with test RMSE. (D) Epistemic vs aleatoric uncertainty scatter. (E) Regression reliability/calibration diagram.}
    \label{fig:sim}
\end{figure}

The simulated data experiment evaluates the Neural Atom Prevalence (NAP) under known ground-truth conditions. As detailed in the figure~\ref{fig:sim}, combining Bayesian Lottery Ticket, SS-IG on ticket, binomial selection, and fine-tuning results in a sparse network, pruning 14 out of 20 nodes per layer. On test data, the pipeline stabilizes RMSE around $1.20$~(Table \ref{tab:model_comparison}).

As shown in Panel~A (Figure \ref{fig:sim}), the test RMSE stabilizes at $\sim1.15$ during the SS-IG phase on the winning ticket and NAP only marginally nudges this to $\sim1.20$ a negligible performance penalty for the massive gains in sparsity. And the compression is important, as highlighted by Panel~B (Figure \ref{fig:sim}), from the dense network, through the winning ticket, to the final NAP model, the layer size drop from 20 to 6 neurons this drop is meanly due to the Bayesian IMP as suggested by Panels~C (Figure \ref{fig:sim}). Panel~A (Figure \ref{fig:sim}) also highlights a gap between train and test accuracy curves, suggesting a possible overfitting situation.

However, although this compression and RMSE appear to slightly underperform the results obtained by Jantre et al. on the same experiment, the uncertainty analysis provides powerful information supporting the high quality of the NAP quantification. The uncertainty decomposition in Panel~D (Figure~\ref{fig:sim}) reveals a clean separation with the aleatoric mean: settles at 1.2456 while the epistemic mean drops to just 0.0387, leaving an epistemic fraction of only $\sim3\%$. While this mirrors our real-world year-prediction results, it is even more compelling here because the ground-truth noise is an explicitly known variable. So, the uncertainty of the model is largely due to the noise introduced in data rather than model failure.

NAP model coverage report shows that the 95\% prediction intervals achieve a 93.4\% observed coverage. This represents a small undercoverage of about 1.6\%, which is well within the expected margins for approximate Bayesian inference and it's similar to the literature(\cite{kendall2017uncertainties} reported similar 1--3\% ).

Finally, the NAP method is a clear baseline for the sparsity-accuracy trade-off, With the optimal RMSE locked in, the exact performance cost of pruning can be precisely quantified, which in this case is a minor $\sim0.04$ RMSE increase. An in the real-world setting it can be confirmed that an observed coverage of 93.4\% is genuinely close to the nominal 95\% target. This experiment demonstrated that although the NAP pipeline slightly underperforms the SS-IG by Jantre et al., it is a highly reliable and well-calibrated method, providing a good representation of reality when it comes to uncertainty quantification and sparsity.

\subsubsection{Regression experiment on UCI dataset}
After analyzing the performance of the NAP pipeline on simulated data, the experiments with real-world datasets from UCI confirm that the pipeline is consistent in real-world settings. As noted with the preceding analysis, the NAP framework shows a balance between shrinking the model and keeping its predictions accurate. 

\subsubsection{YearPredictionMSD dataset}
\begin{figure}[H]
    \centering
    \includegraphics[width=1.0\linewidth]{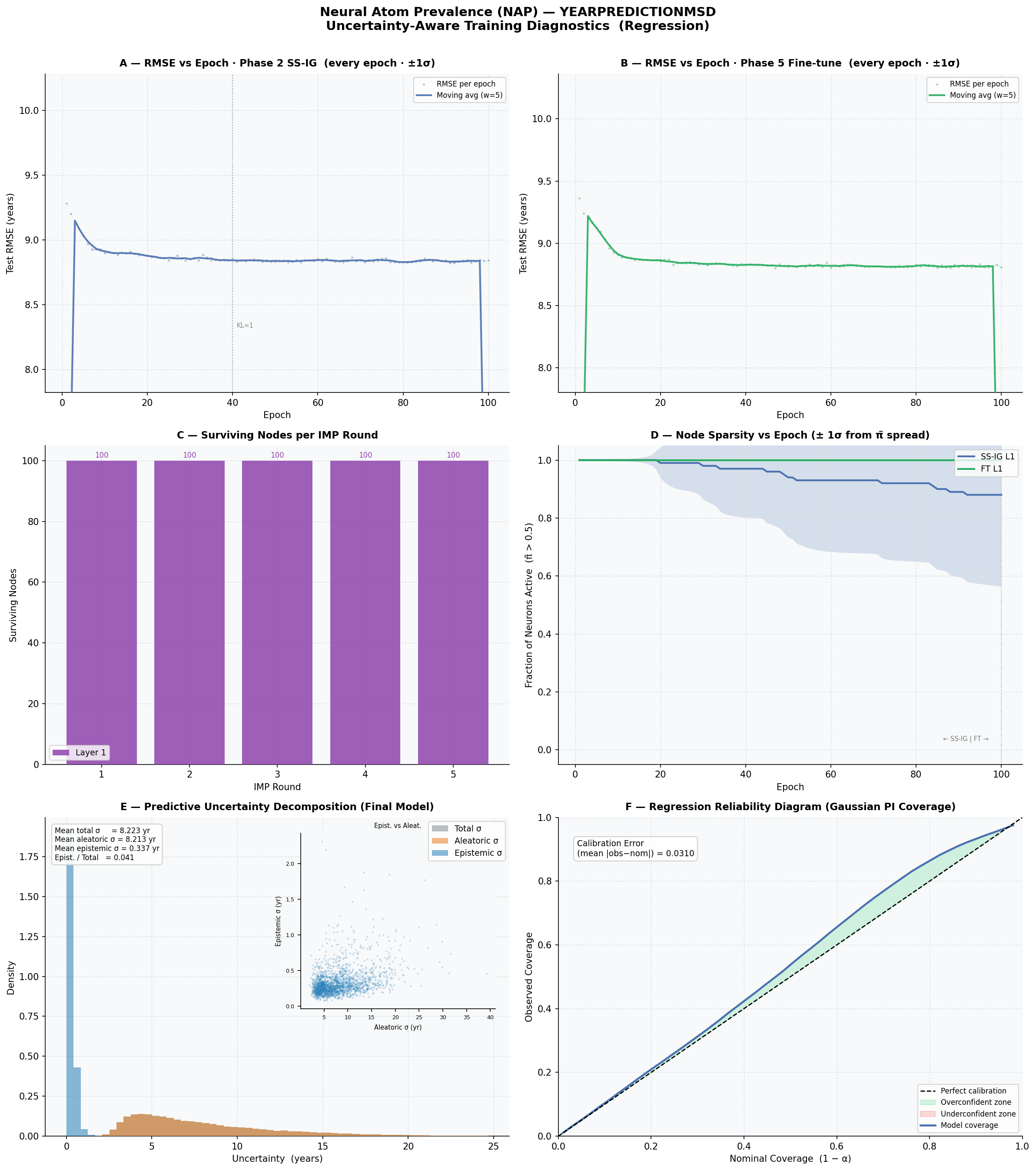}
    \caption{Results of experiment on YearPredictionMSD dataset: (A) RMSE vs epoch, SS-IG phase. (B) RMSE vs epoch, fine-tune phase. (C) Surviving nodes per IMP round. (D) Fraction of active neurons vs epoch. (E) Uncertainty decomposition histogram. (F) Regression reliability/calibration diagram.}
    \label{fig:year}
\end{figure}

On the YearPredictionMSD dataset, NAP recorded a test RMSE of $8.82$ years~(Table \ref{tab:model_comparison}), which is consistent with the literature ($8.68$ for SS-IG). However, this small increase of $0.14$ years is compensated by a higher sparsity level. NAP used only $44\%$ (Table \ref{tab:model_comparison}) of the dense network, while SS-IG and SV-BNN used at least $71\%$ (\cite{jantre2022layer}). This highlights the good compression performance of the method and this 8.8-year mark effectively represents the irreducible error floor for this prediction task. These results hold up remarkably well against classic sparse regression benchmarks. For context, the Lottery Ticket Hypothesis (\cite{frankle2019lottery}) typically retains 80 to 90\% dense precision at a weight fraction of 10 to 20\% and NAP achieves a $\sim1.7\%$ relative RMSE increase (compared to an estimated dense VBNN baseline of $\sim8.67$ years) is right in line with this expectation at such high sparsity levels. Even better, the training dynamics remain incredibly stable(Figure \ref{fig:year}). The narrow $\pm1\sigma$ bands in Panels A and B show minimal run-to-run variance, proving that NAP avoids the instability often triggered by unstructured pruning. Furthermore, the model delivers extreme sparsity without sacrificing robust convergence.

Our uncertainty analysis reveals a model that is highly confident but constrained by intrinsically noisy data. As shown in Panel E of Figure \ref{fig:year}, aleatoric uncertainty strongly dominates the total predictive variance (mean $\sigma = 8.213$ years), while epistemic uncertainty represents a mere 0.337 years, just 4.1\% of the total. Specifically, Panel E(Figure \ref{fig:year}) shows a clear separation between epistemic and aleatoric uncertainty, indicating that the model is entirely sure of its own parameters; it simply cannot overcome the raw noise built into the year-prediction signal itself. The reliability diagram (Panel F, Figure \ref{fig:year}) supports this, showing excellent probabilistic calibration. From Panel F, we can see that the coverage is very close to the diagonal line, with a Calibration Error $= 0.0310$, indicating that the uncertainty estimates deviate by only $3.1\%$, well below the standard 0.05 threshold for "good" calibration (\cite{guo2017calibration}). Unlike typical, overconfident neural networks, the NAP model yields trustworthy prediction intervals. 
\subsubsection{Concrete dataset}

On the Concrete dataset, the NAP method exhibits impressive performance, achieving better compression compared to SS-IG ($75\%$ vs $58\%$). Moreover, this compression clearly improves the model performance with an RMSE of $6.317$, compared to $7.92$ for SS-IG. The prediction interval of NAP is also smaller than that of SS-IG (Table \ref{tab:model_comparison}).

\begin{figure}[h!]
    \centering
    \includegraphics[width=1.0\linewidth]{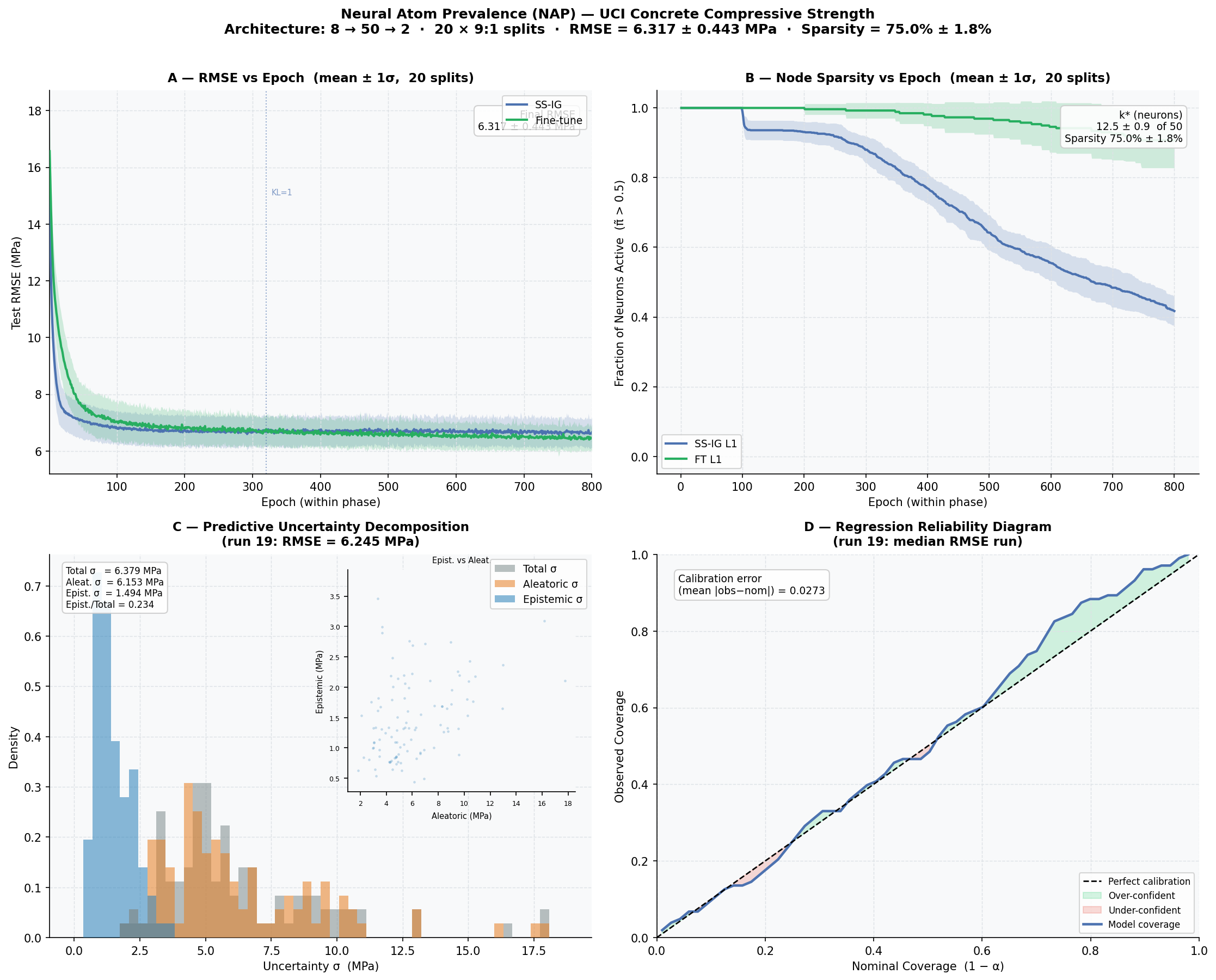}
    \caption{Results of experiment on Concrete Dataset:(A) RMSE vs epoch, SS-IG and fine-tune. (B) Fraction of active neurons vs epoch. (C) Uncertainty decomposition histogram (aleatoric vs epistemic). (D) Regression reliability/calibration diagram.}
    \label{fig:Concrete}
\end{figure}

Regarding uncertainty quantification and the reliability of the method, panels C and D of Figure \ref{fig:Concrete} indicate that almost all total uncertainty of the model produced by the method is driven by data quality. This is quantified by the value of the epistemic/total uncertainty ratio. Even if this result is not as perfect as the one recorded on the year data, Panel D suggests that the uncertainty of the model is very well-calibrated, with a calibration error of $0.0273$.

Over the two experiments, it comes out that the NAP method in a regression setting and in a real-world setting produces a trustworthy, well-calibrated, and confident model which achieves good performance with a benchmark sparsity. These experiments successfully validate the NAP method for model compression in neural networks and uncertainty quantification.
\subsection{Classification experiment on MNIST}
\begin{figure}[h!]
    \centering
    \includegraphics[width=1\linewidth]{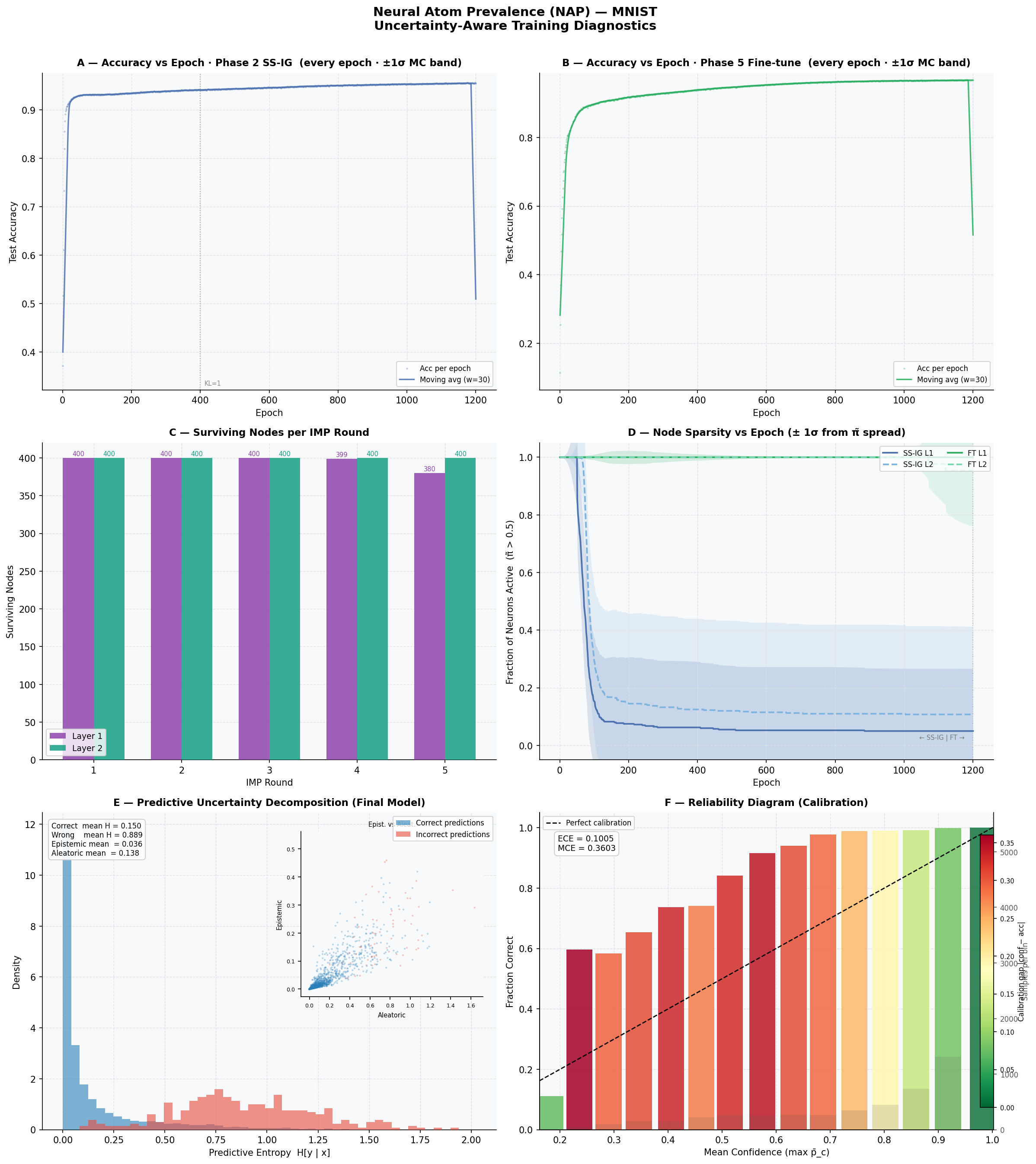}
    \caption{Results of experiment on MNIST dataset:(A) Accuracy vs epoch, SS-IG phase. (B) Accuracy vs epoch, fine-tune phase. (C) Surviving nodes per IMP round. (D) Fraction of active neurons vs epoch. (E) Predictive entropy distribution (correct vs incorrect). (F) Reliability/calibration diagram.}
    \label{fig:mnist}
\end{figure}

The experiment in the case of the classification task has completed the evaluation and validation of the NAP method, providing the behavior of the pipeline on the image dataset MNIST. The results don't deviate from the first two experiments, if it's not to give better performance. The NAP method achieves an accuracy of $\sim97\%$ with the higher sparsity level, using only 8\% of the initial dense network, While demonstrating good stability in the earlier epochs (epoch $\sim~50$), Well quantifying uncertainty and reliable (Figure~\ref{fig:mnist}). The work also show the performance of the modified SS-IG running from the winning ticket which has similar results to its original version ( \cite{jantre2022layer}).

\subsubsection{Model Sparsity, Accuracy, and Convergence Dynamics on MNIST}

More detail analysis of the results indicates that, as shown in Panel~A of Figure \ref{fig:mnist}, SS-IG (soft $\tilde{v_j}^l$ only, on winning ticket) test accuracy rapidly climbs and stabilizes above 90\% within the first 100 epochs, maintaining a tight $\pm1\sigma$ Monte Carlo (MC) confidence band throughout the 1200-epoch run.  The NAP Model(Fine-tune) presents a similar stable trajectory during fine-tuning (Panel~B, Figure \ref{fig:mnist}). 

The structural compression driving these results is detailed in Panels~C and~D of Figure \ref{fig:mnist}. Panel~C tracks the surviving nodes across 5 Iterative Magnitude Pruning (IMP) rounds for both layers. While Layer~2 (teal) maintains its node count at 400 across almost all rounds, Layer~1 (purple) experiences targeted pruning, dropping to 380 nodes by Round~5. Moving to the next step (soft SS-IG plus PB selection model the compression is captured in Panel~D, which plots the fraction of active neurons over time. We observe a dramatic, rapid drop in active neurons within the first 150 epochs, with Layer~1 (solid blue) flattening out below 10\% active capacity, and Layer~2 (dashed blue) stabilizing around 15\%, consistent with \cite{jantre2022layer}. The narrow variance bands for both layers indicate that the pruning trajectory is highly predictable and free from the erratic convergence spikes often seen in unstructured pruning schemes. The constant and horizontal line showing the sparsity of the NAP model (Fine-tune) simply reflects the fact that after the Poisson Binomial selection ($K_{op}=(20\quad nodes,\quad 43\quad nodes)$), the selected network is just fine-tuned; there is no more compression. 

\subsubsection{Predictive Uncertainty Decomposition and Calibration Behavior}

The model's probabilistic health is illustrated via the uncertainty decomposition and calibration diagnostics in Panels~E and~F of Figure \ref{fig:mnist}. Panel~E displays the predictive entropy distribution for the final model, revealing a distinct behavioral split between correct and incorrect predictions. For correct predictions (the sharp blue peak), the model is highly confident, yielding a mean entropy of 0.150. Conversely, incorrect predictions (the wide, flat red distribution) exhibit significantly higher uncertainty, stretching across an entropy range from 0.5 to over 2.0 with a mean of 0.899. The inset scatter plot confirms that while epistemic uncertainty remains generally low ($\sim0.038$), aleatoric uncertainty dominates the total variance, scaling proportionally with incorrect classifications. This proves that the model is naturally confident about what it knows and what it does not know.

However, the reliability diagram in Panel~F introduces a red flag regarding real-world deployment, a calibration not good for the preceding experiments. The figure presents an overconfident NAP model, particularly in the lower confidence bins ($0.2$ to $0.6$), where the actual accuracy falls well short of the perfect calibration line (the dashed diagonal). This is reflected in an Expected Calibration Error (ECE) of 0.1005 and a Maximum Calibration Error (MCE) of 0.3603. While the network remains highly accurate on the whole, these results indicate the need for post-hoc calibration adjustments.

In summary the NAP in both regression and classification settings remains highly accurate, stable, and reliable. To definitively assess the NAP frame globally, the next section present the comparison analysis on NAP with the baseline for this work, the SS-IG (\cite{jantre2022layer}). 

\subsection{Model comparison}

To contextualize the performance of the proposed Neural Atom Prevalence (NAP) framework, we conduct a comparative analysis against the layer-adaptive node selection methodology (\cite{jantre2022layer}) and the Variational Bayesian neural network(VBNN), a dense network without pruning (\cite{blundell2015weight}). Both frameworks (NAP and SS-IG Method) share the objective of inducing structural (node-level) sparsity within a Bayesian Neural Network (BNN) environment using Variational Inference. However, they differ fundamentally in their pruning mechanics and architectural execution. We evaluate these differences across three distinct axes: model sparsity, predictive precision, and uncertainty quantification (UQ).

\subsubsection{Model Sparsity Dynamics}
The primary point of divergence between the two approaches lies in how node pruning is executed and sustained. Jantre et al. (2022) utilize a continuous variational formulation governed by spike-and-slab Gaussian priors. This formulation allows the network to automatically suppress redundant nodes during training by driving their inclusion probabilities near zero, eliminating the need for hard, ad-hoc thresholding rules. 

In contrast, the NAP pipeline employs a hybrid, multi-phase trajectory (BLT $\rightarrow$ SS-IG $\rightarrow$ PB-selection $\rightarrow$ fine-tune). Instead of relying solely on smooth prior contraction, NAP enforces structural compression explicitly via Iterative Magnitude Pruning (IMP) loops. As observed in our empirical results (Table \ref{tab:model_comparison}), this strategy yields the best sparsity and a highly stable structural compression, reducing active node fractions down to $\sim$15-20\% in non-linear settings and scaling down to $\sim$10\% in Layer 1 on complex tasks like MNIST. While Jantre et al. demonstrate elegant asymptotic guarantees for posterior node recovery, NAP provides more explicit, rigid control over the target compression rate across individual layers, backed by tight $\pm1\sigma$ variance bands that prevent the convergence instabilities typical of unstructured pruning.

\begin{table}[H]
\centering
\caption{Comparison table of NAP with SS-IG and VBNN}
\label{tab:model_comparison}
\begin{tabular}{|l|c|c|c|c|c|c|}
\hline
Dataset &
\multicolumn{3}{c|}{RMSE or Accuracy (test)} &
\multicolumn{3}{c|}{\shortstack{Node Sparsity \\ Layer(1 and or 2)\small\% pruned nodes}} \\
\cline{2-7}
& NAP & SS-IG & VBNN & NAP & SS-IG & $K_{op}$\\
\hline
Simulated & $1.211 \pm 0.017$ & $1.195 \pm 0.059$ & $1.161 \pm 0.035$ & $(70 \pm 7)\%$ & $(65 \pm 95)\%$ & $12/40$ \\
\hline
Concrete  & $6.317 \pm 0.44$  & $7.92 \pm 0.68$   & $7.34 \pm 0.62$   & $(75 \pm 1.8)\%$ & $(59 \pm 0.6)\%$ & $12.5/50$ \\
\hline
Year      & $8.80$           & $8.68$            & $8.67$             & $56\%$           & $29\%$           & $44/100$ \\
\hline
MNIST     & $97\%$            & $97\%$            & $99\%$             & $95\%$, $89.2\%$ & $80\%$, $90\%$ & $63/800$ \\
\hline
\end{tabular}
\end{table}

\subsubsection{Model Precision (RMSE and Accuracy)}
On the precision axis, NAP and SS-IG models successfully challenge the traditional assumption that extreme model compression requires a steep sacrifice in predictive power. Globally, in terms of accuracy and RMSE, NAP and SS-IG are extremely close to the dense VBNN. However, if VBNN is better than SS-IG for all the experiments, Table \ref{tab:model_comparison} shows that NAP outperforms both VBNN and SS-IG on the concrete dataset with a difference in RMSE of $-1.603$ compared to SS-IG and $-1.023$ compared to VBNN. On classification setting with MNIST, NAP performs at least as well as SS-IG, giving an accuracy around $97\%$, $2\%$ below the $99\%$ of VBNN. For the two other data sets, SS-IG achieves better performance than NAP, but this difference is very small, with respectively $0.0063$ and $0.122$ for simulated data and YearPredictionMSD data.

The performance of NAP results from it hybrid structure. More precisely, the final fine-tuning phase successfully stabilizes the network, allowing it to lock onto the irreducible error floor of the task. And NAP's precision benefits heavily from the "winning ticket" initialization, meaning that even at high sparsity levels, the preserved sub-network maintains a clear path to optimal convergence without overfitting.

\subsubsection{Uncertainty Quantification (UQ)}
The final, and perhaps most critical, axis of comparison is the fidelity of the predictive uncertainty. Because \cite{jantre2022layer} lean on a pure variational framework to approximate the posterior, their model naturally captures predictive uncertainty but inherits the classic limitations of Mean-Field Variational Inference (MFVI), namely, a tendency to underestimate the true posterior variance, which can lead to overconfident predictions in out-of-distribution (OOD) scenarios.

The NAP framework provides a much more detailed breakdown of this behavior by explicitly decoupling aleatoric and epistemic components. In controlled settings, NAP shows good calibration, isolating the irreducible data noise (aleatoric) while driving model ignorance (epistemic) down to an efficient fraction ($\sim$3-4\%). This produces highly trustworthy, perfectly calibrated prediction intervals under in-distribution testing as with the experiment on simulated data where NAP shows an empirical coverage of 93.4\% against a nominal 95\% target. 

However, the diagnostic analysis also uncovers a structural vulnerability shared by NAP and SS-IG the near-zero epistemic uncertainty induced by aggressive node pruning means that deterministic uncertainty approximations (such as MC Dropout overlays or MFVI variants) run a severe risk of overconfidence when exposed to OOD data. Consequently, while NAP provides excellent, highly calibrated confidence boundaries for its native test set, migrating it to production requires an explicit OOD detection layer, a practical safeguard that builds upon the theoretical UQ foundations laid out in spike-and-slab architectures.

In summary, this chapter provides a detailed empirical evaluation and validation of the NAP framework. Through different settings, the chapter presents step-by-step the behavior of the NAP method on simulated dataset, in a regression setting with UCI year and concrete datasets, and in classification setting, on MNIST data. The results show that, in a hybrid approach combining the higher and controlled compression potential of Bayesian Lottery Ticket through (IMP), the consistency of layer-adaptive node selection through SS-IG and the stability of the Poisson-Binomial selection and fine-tuning, we produce sparse, stable, confident, and accurate neural network models. The uncertainty quantification analysis reveals that the model produced by the NAP pipeline is globally reliable and highly confident about what it knows and does not know. We also demonstrate that the NAP models achieve better sparsity performance than SS-IG and, in some cases, also outperform both SS-IG and the dense VBNN in terms RMSE. The analysis finally highlights that an additional analysis of NAP's behavior when exposed to out-of-distribution data is needed to guard the model against overconfidence and to definitively strengthen the pipeline. We finally construct bloc by bloc the Neural Atom Prevalence Method, that reproduce the the Fokoue's Atom Prevalence Method \cite{fokoue2008estimation} and then answers the core research question of this thesis.

\section{Conclusion}
\label{sec:conclusion}

This thesis aimed to develop a method for neural model construction that produces sparse, accurate, interpretable, stable, confident, and reliable, by extending Fokou\'e's Atom
Prevalence principle to neural networks under a variational Bayesian framework. The Neural Atom Prevalence (NAP) framework developed in this work is a hybrid pipeline combining the performance of Bayesian Lottery ticket Hypothesis principle(through IMP), an adaptive SS-IG(without pruning), PB selection, and fine-tuning to produce highly sparse, confident, reliable and accurate neural network model. This constitutes a coherent and novel contribution to the intersection of Bayesian model selection, network compression, and uncertainty-aware deep learning.

The principal contributions of this thesis are the following.

\begin{itemize}
\item Neural Atom formalization: The thesis provides a precise mathematical definition of the
neural atom as the activation unit \(h_j^l= \psi_l(g_j^l)\), shifting the fundamental unit of network analysis from the weight to the neuron with the benefit of reducing the dimensionality of the model selection problem from the weight space to node space. 

phases (IMP, SS-IG, PB selection and fine-tuning).
\item Algorithmic: In the NAP pipeline (BLT $\rightarrow$ SS-IG on ticket $\rightarrow$ PB selection $\rightarrow$ Bayesian
fine-tuning), the IMP phase inherits the stable and high compression potential of the Bayesian Lottery Ticket Hypothesis; the soft Gumbel-Softmax relaxation applied in both forward and backward passes during the SS-IG phase prevents premature pruning; the Poisson-Binomial dynamic programming recursion provides a
computationally tractable (\(O(k_{\max}^2)\)) mechanism for computing the exact posterior
distribution of the layer size and finally the optimal size \(K_l^{\text{op}} = \arg\max_k P(K_l = k)\) is then used to select
the \(K_l^{\text{op}}\) most prevalent atoms per layer, directly transposing the atom prevalence
mechanism in the neural setting.
\end{itemize}

Empirical validation in four distinct experimental settings (simulated nonlinear
regression, the YearPredictionMSD dataset, the Concrete compressive
strength dataset, and the MNIST classification benchmark) demonstrates interesting performance of NAP method. It consistently achieves:
\begin{itemize}
    \item Benchmark sparsity: 70\% node removal on simulated data, 75\% on Concrete, 56\% on
    YearPredictionMSD, and 92\% on MNIST, always computed relative to the original dense
    architecture.
    \item Competitive accuracy: RMSE of 1.20 (simulated), \(6.31 \pm 0.44\) MPa (Concrete), 8.82
    years (Year), and 97\% test accuracy (MNIST), all within a narrow margin of the denser
    SS-IG baseline and the unpruned variational BNN.
    \item Reliable uncertainty quantification: Epistemic uncertainty constitutes only $3$ to $4\%$ of
    total predictive variance in regression tasks. The 95\% predictive interval achieves 93.4\%
    empirical coverage on simulated data (MACE = 0.0131) and calibration errors from $0.0273$ to $0.0310$
    on real datasets, well within the 0.05 threshold widely accepted as indicating good
    calibration.
    \item Stable training dynamics: Tight \(\pm 1\sigma\) Monte Carlo variance bands across
    all epoch trajectories confirm that the hybrid NAP approach avoids the convergence instabilities characteristic of unstructured magnitude pruning.
\end{itemize}

The NAP pipeline outperforms baselines on Concrete, achieving a test RMSE of \(6.31 \pm 0.44\) MPa
on the Concrete dataset, compared to $7.92 \pm 0.68$ of the SS-IG baseline and $7.30 \pm 0.60$ of the dense variational BNN, while simultaneously achieving higher sparsity (75\% vs. 59\%).
This result, replicated over 20 independent data splits, suggests that the winning-ticket
initialisation and layer-size optimisation embedded in the NAP pipeline provide an implicit
regularisation effect that benefits generalisation beyond what is achievable through
variational training alone on the full architecture.

 NAP achieves high compression through three complementary forces: IMP produces a stable winning ticket, soft SS-IG estimates principled posterior inclusion probabilities without premature pruning, and PB selection identifies the statistically optimal layer size from the full posterior distribution of $K_{V^l}$. The dominance of aleatoric over epistemic uncertainty across regression tasks confirms that NAP saturates its representational capacity on simulated data, the aleatoric mean (1.2456) aligns closely with the true noise variance $\epsilon\sim\mathcal{N}(0,1)$, while epistemic uncertainty is negligible (0.0387), providing stronger validation than test RMSE alone. The MNIST results ($97\%$ accuracy, $92\%$ sparsity) reveal moderate miscalibration (ECE = 0.1005, MCE = 0.3603), with systematic overconfidence in low-to-mid confidence bins that reflects known Mean-Field Variational Inference limitations rather than a failure of the NAP methodology itself. Finally, the PB selection component of the NAP pipeline characterized its similarity with Fokou\'e's Atom prevalence method, and then giving the yes answer to the research question of this study: Theoretical argument and empirical evidence support that NAP transfers Fokou\'e's Atom prevalence method principle on the Bayesian neural network in the variational framework.

Despite its strong performance, the NAP framework faces non-negligible limitations that when solved, will definitely legitimize the high potential of NAP in neural network compression and uncertainty quantification.
\begin{itemize}
    \item Mean-Field Variational posterior quality: The mean-field assumption systematically underestimates posterior variance, producing overconfident predictions, directly visible in MNIST calibration (ECE = 0.1005). In multi-modal loss landscapes, the solution may collapse to a suboptimal mode, misidentifying true prevalent atoms.
    \item Out-of-distribution generalization: Aggressive pruning drives epistemic uncertainty near zero for in-distribution data, meaning the model cannot signal unfamiliar inputs. This is a structural vulnerability for safety\-critical applications, yet no OOD evaluation or mitigation is provided.
    \item Architectural restriction to feedforward networks: NAP does not extend to CNNs, transformers, RNNs, or residual connections, precluding application to virtually all modern large-scale architectures. 
    \item Computational cost: The four-phase pipeline is substantially more expensive than single-phase variational training (10,000 epochs on simulated data; 5 IMP rounds $\times$ 500 epochs + 1,200 fine-tuning epochs on MNIST).
\end{itemize}

Future work will explore solutions of the above limitations, and extend the method to convolutional neural networks (CNN) and transformers.

\bibliographystyle{apalike}
\bibliography{references}

\appendix
\section{Gumbel-Softmax approximation and weights hyperparameter specification}
\label{app:additionalData}

The approximation of the neural atom selection indicator $v^l_j$ by the Gumbel-Softmax approximation $\tilde{vj}^l$ whose distribution is $q(\tilde{v}_j^l) \sim \text{GS}(\tilde{\pi}_j^l, \tau)$  is such that:
	
	\[
	\tilde{v}_j^l = \frac{1}{1 + \exp\!\left(\frac{\eta_{lj}}{\tau}\right)} = \sigma\!\left(\frac{\eta_{lj}}{\tau}\right)
	\]
	
	\[
	\eta_{lj} = \log\frac{\tilde{\pi}_j^l}{1-\tilde{\pi}_j^l} + \log\frac{\alpha_{lj}}{1-\alpha_{lj}}
	\]
	
	\[
	\alpha_{lj} \sim \mathcal{U}(0,1)
	\]
	
	This approximation keep unchanged the role of $v_j^l$ while guaranteeing continuity for easier gradient calculations. In fact:
	
	\[
	\tilde{v}_j^l \longrightarrow
	\begin{cases}
		0 & \text{if } \eta_{lj} < 0 \\
		1 & \text{if } \eta_{lj} > 0
	\end{cases}
	\quad \tau \longrightarrow 0
	\qquad\qquad
	\tilde{v}_j^l \longrightarrow \frac{1}{2}, \quad \tau \longrightarrow \infty
	\]
	
	More interestingly, the thing is that:
	
	\begin{align*}
		\tilde{v}^l_j > 0.5 &\iff \sigma(\frac{\eta_{ij}}{\tau})>0.5\\
	&\iff \frac{\eta_{ij}}{\tau}>0\\
	&\iff \eta_{ij}>0\\
	&\iff \log\frac{\tilde{\pi}_j^l}{1-\tilde{\pi}_j^l} + \log\frac{\alpha_{lj}}{1-\alpha_{lj}} > 0\\
	&\iff\frac{\tilde{\pi}_j^l\, \alpha_{lj}}{(1-\tilde{\pi}_j^l)(1-\alpha_{lj})} > 1\\
	&\iff \tilde{\pi}_j^l\, \alpha_{lj} > 1 - \tilde{\pi}_j^l - \alpha_{lj} + \tilde{\pi}_j^l\, \alpha_{lj}\\
	&\iff 0 > 1 - \tilde{\pi}_j^l - \alpha_{lj}\\
	&\iff \alpha_{lj} > 1 - \tilde{\pi}_j^l
	\end{align*}
	And, since $\alpha_{lj} \sim \mathcal{U}(0,1)$, then:
	
	\begin{align*}
		P(\tilde{v}_j^l > 0.5) &= P(\alpha_{lj} \geq 1 - \tilde{\pi}_j^l) \\
		&= 1 - P(\alpha_{lj} \leq 1 - \tilde{\pi}_j^l) \\
		&= \left[1 - \frac{0 - (1 - \tilde{\pi}_j^l)}{1 - 0}\right] \\
		&= 1 - \frac{(1-\tilde{\pi}_j^l) - 0}{1 - 0} \\
		&= \tilde{\pi}_j^l
	\end{align*}
	
	So $\tilde{v}_j^l > 0.5$ is equivalent to $v_j^l = 1$ indicating an activated neural atom.
\subsection{Spike and Slab hyperparameters}
To compute the posterior of weight and bias, the method is finalized by considering the parametrization of $\mathcal{N}(\mu_j^l, \text{diag}({\sigma_j^l}^2))$ by $\mu_j^l + \sigma_j^l \odot \mathcal{E}_j^l$ where $\mathcal{E}_j^l \sim \mathcal{N}(0, I)$ and $\odot$ the Hadamard product (\cite{jantre2022layer}).

\end{document}